\RequirePackage[loading]{tracefnt}
\documentclass[12pt,a4paper]{article}

\usepackage{graphicx}

\PassOptionsToPackage{numbers, compress}{natbib}

\usepackage[margin=1.0in]{geometry}
\usepackage[square, numbers]{natbib}
\usepackage[hidelinks, hypertexnames=false]{hyperref}      
\usepackage{url}       
\usepackage{booktabs}     
\usepackage{amsfonts}      
\usepackage{nicefrac}  
\usepackage{microtype}      
\usepackage{xcolor}      

\usepackage{enumitem}
\usepackage{mathtools, amsthm}
\usepackage[utf8]{inputenc}
\usepackage{float}
\usepackage{subcaption}
\usepackage{listings}
\usepackage{amsmath}
\usepackage{newtxmath}

\usepackage{multirow}

\newcommand{\norm}[1]{\left\lVert#1\right\rVert}

\title{\textbf{Scalable Signature-Based Distribution Regression via Reference Sets}}

\author{\vspace{-0.35em} \textbf{Andrew Alden} \\\vspace{-0.35em}
  \small Department of Informatics\\\vspace{-0.35em}
  \small King's College London\\\vspace{-0.35em}
  \small \texttt{andrew.alden@kcl.ac.uk}
  \and
  \vspace{-0.35em} \textbf{Carmine Ventre} \\\vspace{-0.35em}
  \small Department of Informatics\\\vspace{-0.35em}
  \small King's College London \\\vspace{-0.35em}
  \small \texttt{carmine.ventre@kcl.ac.uk} \\\vspace{-0.35em}
  \and
  \vspace{-0.35em} \textbf{Blanka Horvath} \\\vspace{-0.35em}
  \small Department of Mathematics \\\vspace{-0.35em}
  \small University of Oxford \\\vspace{-0.35em}
  \small \texttt{blanka.horvath@maths.ox.ac.uk}
}
\date{}

\begin{document}

\maketitle

\begin{abstract}
Distribution Regression (DR) on stochastic processes describes the learning task of regression on collections of time series. Path signatures, a technique prevalent in stochastic analysis, have been used to solve the DR problem. Recent works have demonstrated the ability of such solutions to leverage the information encoded in paths via signature-based features. However, current state of the art DR solutions are memory intensive and incur a high computation cost. This leads to a trade-off between path length and the number of paths considered. This computational bottleneck limits the application to small sample sizes which consequently introduces estimation uncertainty. In this paper, we present a methodology for addressing the above issues; resolving estimation uncertainties whilst also proposing a pipeline that enables us to use DR for a wide variety of learning tasks. Integral to our approach is our novel distance approximator. This allows us to seamlessly apply our methodology across different application domains, sampling rates, and stochastic process dimensions. We show that our model performs well in applications related to estimation theory, quantitative finance, and physical sciences. We demonstrate that our model generalises well, not only to unseen data within a given distribution, but also under unseen regimes (unseen classes of stochastic models).
\end{abstract}

\section{Introduction}\label{sec:intro}

Distribution Regression (DR) \cite{reference_points_paper, dist_reg_paper, dr_1, dr_2} describes the supervised learning task of regressing probability distributions against a target, extending the classical regression framework to the distributional setting. DR has applications in various domains including agriculture \cite{dist_reg_paper}, computer vision \cite{reference_points_paper}, finance \cite{conference_paper, dist_reg_paper, higher_order_kme_neurips}, astrophysics \cite{astrophysics_dr_2, astrophysics_dr_1}, voting behaviour \cite{voting_dr_1, voting_dr_2}, and many more. A significant difference between classical regression and DR lies in the availability of the regressor. In classical regression, the regressor is fully observable, while in DR each distribution is only observed through samples. 
In this paper, we generate samples from synthetic distributions to create a synthetic empirical distribution that can serve as an approximation for the true distribution. This empirical distribution then acts as the (proxy) regressor, and the approximation is performed using Kernel Mean Embeddings (KME) \cite{bayesian_kme, kme_review}, a technique by which distributions are represented as points within a corresponding Hilbert space. The accuracy of this approximation is dependent on a number of factors, such as dimensionality of the problem, the variance of the distributions at hand and in particular, the amount of samples available. 
  Initially, results in the DR domain were developed for finite-dimensional distributions, but not (path-valued) data-streams which are often infinite-dimensional.
The DR framework was later extended \cite{dist_reg_paper} to stochastic processes, which is the focus of our work. In our setting, observations stem from data streams (time series, sample paths), where data is recorded as a stream of consecutive, not necessarily equidistant, time steps, sampled from an underlying (unknown) stochastic process.

\section{Related work} 

Since many important statistical learning and Machine Learning (ML) tasks fall under the DR framework, such as the ones listed in Section \ref{sec:intro}, DR has become an active area of research in recent years. As a framework, DR initially focused on regressing distributions of random variables against a scalar target. Previous work focuses on ridge regression techniques (see e.g. \cite{dr_2}) along with kernel learning (see \cite{voting_dr_1}). In \cite{dr_2}, the authors study the optimal DR function when using ridge regression. They prove the consistency of the DR solution under mild assumptions. 
In \cite{voting_dr_1, reference_points_paper}, a Bayesian approach to DR is presented. While \cite{voting_dr_1} incorporate a Bayesian approach by combining kernel ridge regression (KRR) and Gaussian process regression, \cite{reference_points_paper} impose a Bayesian prior over the KMEs to account for the sample variance being propagated through the model. They also use landmark points to perform DR, which is comparable in principle to the solution presented in our work. However, we work in a pathwise setting. The DR framework is closely related to the multi-instance learning paradigm. In multi-instance learning \cite{multi_instance_learning_3, multi_instance_learning_1, multi_instance_learning_4, multi_instance_learning_5, multi_instance_learning_2, deep_sets_paper}, the features are \textit{bags-of-data} which are treated as sets. By considering the data collections as independent samples from a distribution, multi-instance learning is very similar to DR. \newline

DR was then extended to stochastic processes \cite{dist_reg_paper}. \cite{dist_reg_paper} constructed features using the path signature, a transform which converts a time series into a vector representation. The path signature has been shown to complement ML tasks well. It has two main properties which make it ideal for ML tasks \cite{vae_paper, sig_ml, nrde_paper, chinese_writing} and DR in particular; the expected signature characterises the law of the stochastic process (under certain assumptions) \cite{unique_sig_dist} and a non-linear function of a path can be approximated arbitrarily well by a linear function of the terms in the signature \cite{sig_ml_2}. When working with stochastic processes, an important factor is the filtration, which represents the information observable (a.k.a., measurable) at any given time. The possibility of capturing information flow (via the filtration) within the ML methodology improves prediction accuracy on tasks where these variables are essential, see \cite{conference_paper, american_option_dr, higher_order_kme_neurips}. To embed the filtration within the DR framework, higher-order KMEs have been developed in \cite{higher_order_kme_neurips}. Using higher-order signatures in ML models has a number of advantages as laid out in \cite{conference_paper, american_option_dr, higher_order_kme_neurips}, but it is computationally challenging and memory intensive. As a result of intense memory requirements, current state of the art DR approaches face a trade-off between; (1) the number of samples considered, (2) the dimension of the stochastic process, and (3) the length of the time series. In practice, as a result of the above, current DR approaches are limited to tasks involving small sample sizes and small datasets. Small sample sizes are known to result in a noisy approximation of the empirical distribution, leading to ambiguity about the true underlying distribution, which is one of the challenges we successfully address in this work.

\section{Contributions}

In this work, we combine two lines of research that have already yielded remarkable results. Namely, we perform DR on path-valued random variables via higher-rank signatures and we use landmark points to address the aforementioned computational issues relating to the (otherwise very powerful) higher-order DR framework \cite{higher_order_kme_neurips}. We present Signature-Powered Estimation using Empirical Distances to Reference Sets (SPEEDRS), a distance-based approach where each stochastic process is evaluated by calculating distances to fixed reference sets containing samples from stochastic processes. These distances are calculated using our novel distance approximator. By using our distance approximator, we alleviate a large portion of the computational overhead from the methodologies used in DR with (higher-rank) signatures. An added advantage is that with this solution, we also gain further model features. Having a pathwise distance approximator allows us to seamlessly apply our methodology across different application domains, sampling rates, stochastic process dimensions, and unseen regimes. These were verified experimentally across three different ML tasks, with strong results across a number of dimensions (discussed in Section \ref{sec:exp}). Our methodology directly tackles the following issues; resolving estimation uncertainties whilst also proposing a pipeline that enables us to apply DR to a broad range of application domains and datasets.

\section{Theoretical background} \label{sec:background}

\paragraph{Notation} 
Let \(\mathbb{T} \coloneqq \left[0, T\right]\) denote a continuous time index and \(\mathcal{X} \left(\mathcal{K} \right) \coloneqq \{\pmb{x} \colon \mathbb{T} \to \mathcal{K} \}\) denote the compact set of continuous, piecewise linear paths with values in \(\mathcal{K}\) obtained by linearly interpolating a discrete time path \(\pmb{x} = \left(\pmb{x}_{t}\right)_{t \in \mathbb{T}^{N}}\). Let \(\pmb{x}_{t}^{k}\) denote the value of the \(k^{\text{th}}\) dimension of the path at time $t$. Also, let \(\mathbb{X} \coloneqq \left( \Omega, \mathcal{F}, \mathbb{F}, \mathbb{P}, \pmb{X} \right)\) be a filtered process \cite{tuple_defn} where \(\pmb{X} \coloneqq \left( \pmb{X}_{t} \right)_{t \in \mathbb{T}}\) is a \(\mathcal{K}\)-valued \(\mathbb{F}\)-adapted process defined on \(\Omega\).

\subsection{Problem Formulation} \label{sec:dr_problem}

Consider \(M\) pairs \(\left( \mathbb{X}^{i}, y^{i} \right)\) where each pair consists of a \(d\)-dimensional stochastic process \(\mathbb{X}^{i}\) modelling the dynamics of a system (such as weather, stock prices) and a scalar target \(y^{i}\). Let $\pmb{x}^{i, k}_{t_{j}}$ denote the $k^{\text{th}}$ sample of the $i^{\text{th}}$ stochastic process observed at time $t_{j}$. Augment the time series with a time component as follows;
\begin{equation*}
\left\{ \pmb{x}^{i, k} = \left\{ \left(t_{0}, \pmb{x}_{t_{0}}^{i, k} \right), \left(t_{1}, \pmb{x}_{t_{1}}^{i, k }\right), \cdots, \left(t_{l_{i, k}}, \pmb{x}_{t_{l_{i, k}}}^{i, k} \right) \right\}  \right\}_{k=1}^{N_{i}} \sim \mathbb{X}^{i}.
\end{equation*}
The discrete paths $\pmb{x}^{i, k}$ can be embedded in a Lipschitz continuous path \(\tilde{\pmb{x}}^{i, k} \colon \left[t_{0}, t_{l_{i, k}} \right] \to \mathbb{R}^{d+1}\) using piecewise linear interpolation such that \(\tilde{\pmb{x}}^{i, k} \left(t_{j}\right) = \pmb{x}^{i, k} \left(t_{j}\right)\) for all \(j = t_{0}, \cdots, t_{l_{i, k}}\). The empirical probability distribution of $\mathbb{X}^{i}$ is then given by \(\delta^{i} \left( \cdot \right) \coloneqq N_{i}^{-1} \sum_{j=1}^{N_{i}} \delta_{\tilde{\pmb{x}}^{i, k}} \left( \cdot \right)\) where \(\delta_{\tilde{\pmb{x}}^{i, k}} \left( \cdot \right)\) is the Dirac measure\footnote{For any set of paths \(P\), the Dirac measure \(\delta_{\tilde{\pmb{x}}^{i, k}} \left( P \right)\) is \(1\) if \(\tilde{\pmb{x}}^{i, k} \in P\) and \(0\) otherwise.} centered at \(\tilde{\pmb{x}}^{i, k}\). The objective of DR is to construct a function \(F \colon \delta^{i} \to \mathbb{R} \) satisfying some optimality condition. The methodology we use to construct \(F\) is based on distances between stochastic processes. We now describe how these distances are constructed.  

\subsection{Path signature}

The path signature \cite{chen_paper, sig_book, sig_defn_paper_1, sig_defn_paper_2} is a transform which maps a path into a vector representation. The signature \(\mathcal{S}\) of a \(d\)-dimensional path \(\pmb{x} \in \mathcal{X} \left(\mathcal{K}\right)\) is defined as the infinite tensor
\begin{equation*}
    \mathcal{S} \left(\pmb{x}\right) \coloneqq \left(1, \left\{ \mathcal{S} \left(\pmb{x}\right)^{\left(k_{1}\right)} \right\}_{k_{1}=1}^{d},  \left\{ \mathcal{S} \left(\pmb{x}\right)^{\left(k_{1}, k_{2}\right)} \right\}_{k_{1}, k_{2}=1}^{d}, \cdots\right), \textrm{with }  \mathcal{S} \left(\pmb{x}\right)^{\left(k_{1}, \cdots, k_{j}\right)} \coloneqq \int_{R_{j}} d\pmb{x}_{s_{1}}^{k_{1}} \dots d\pmb{x}_{s_{j}}^{k_{j}}
\end{equation*}
\noindent
over the simplex \(R_{j} = \left\{0 < s_{1} \cdots < s_{j} < T\right\}\). Since the paths are piecewise linear and of bounded variation, the integrals are well-defined and are understood in the Riemann-Stieltjes sense. The truncated signature truncated at level \(L\) is the signature containing all terms up until, and including, the collection \(\left\{ \mathcal{S} \left(\pmb{x}\right)^{\left(k_{1}, \cdots, k_{L}\right)}\right\}\). It is denoted by \(\mathcal{S}^{L}\).  \newline

Given two paths $\pmb{x}, \pmb{y}$, the signature kernel \cite{kernel_sequential_data} is defined as \(k_{\mathcal{S}} \left(\pmb{x}, \pmb{y} \right) \coloneqq \langle \mathcal{S} \left(\pmb{x}\right), \mathcal{S} \left(\pmb{y}\right) \rangle\).  \cite{sig_kernel_pde_paper} show that \(k_{\mathcal{S}} \left(\pmb{x}, \pmb{y} \right) = u_{\pmb{x}, \pmb{y}} \left(T, T\right)\) where \(u_{\pmb{x}, \pmb{y}} \colon \left[0, T\right]^{2} \to \mathbb{R}\) is the solution to the hyperbolic partial differential equation \cite[Theorem 2.5]{sig_kernel_pde_paper}

\begin{equation} \label{eqn:goursat_pde}
\frac{\partial^{2} u_{\pmb{x}, \pmb{y}}}{\partial s_{1} \partial s_{2}} = \left\langle \frac{d \pmb{x}}{dt} \left(s_{1}\right), \frac{d \pmb{y}}{dt} \left(s_{2}\right) \right\rangle_{\mathcal{H}} u_{\pmb{x}, \pmb{y}}
\end{equation}
\noindent
with boundary conditions \(u_{\pmb{x}, \pmb{y}} \left(0, \cdot\right) = u_{\pmb{x}, \pmb{y}} \left(\cdot, 0\right) = 1\).

\subsection{Kernel mean embeddings}

Let \(\mathcal{H}_{\mathcal{S}}\) denote the Reproducing Kernel Hilbert Space (RKHS) with \(k_{\mathcal{S}}\) being the reproducing kernel. Let \(\mathbb{X}\) be a filtered stochastic process with values in \(\mathcal{K}\) satisfying the integrability condition \(\mathbb{E}_{\mathbb{P}_{\pmb{X}}} \left[ k_{\mathcal{S}} \left(\pmb{X}, \pmb{X} \right) \right] < \infty\). \(\mathbb{P}_{\mathbf{X}}\) can be embedded in \(\mathcal{H}_{\mathcal{S}}\) using the signature KME \cite{higher_order_kme_neurips}. This is defined as \(\mu_{\pmb{X}}^{1} \coloneqq \mathbb{E}_{\mathbb{P}_{\pmb{X}}} \left[ k_{\mathcal{S}} \left(\pmb{X}, \cdot \right) \right]\). The \(1^{\text{st}}\)-order KME of the conditional law \cite{higher_order_kme_neurips} \(\mathbb{P}_{\pmb{X} | \left[0, t\right]}\) extends the KME and is defined by \(\mu^{1}_{\pmb{X} | \left[0, t\right]} \coloneqq \mathbb{E} \left[ k_{\mathcal{S}} \left(\pmb{X}, \cdot \right) | \mathbb{X}_{\left[0, t\right]} \right] \), where \(\mathbb{X}_{\left[0, t\right]}\) denotes the stochastic process restricted to the times \(\left[0, t\right] \in \mathbb{T}\). By using an expanding conditioning window, we construct the stochastic process \(\mu^{1}_{\pmb{X} | \mathbb{T}} \coloneqq \left( \mu^{1}_{\pmb{X} | \left[0, t\right]}\right)_{t \in \mathbb{T}}\). Similar to the above, the law of \(\mu^{1}_{\pmb{X} | \mathbb{T}}\) can be embedded in a higher-order RKHS \cite{higher_order_kme_neurips}. The \(2^{\text{nd}}\)-order KME is defined as \(\mu^{2}_{\pmb{X}} \coloneqq \mathbb{E} \left[k_{\mathcal{S}} \left(\mu^{1}_{\pmb{X} | \mathbb{T}}~, \cdot\right)\right]\).

\subsection{The Maximum Mean Discrepancy}

The KMEs map the distribution to a unique point in the RKHS. By computing the distance between KMEs in the Hilbert Space, we can obtain a representative distance between distributions \cite{first_order_mmd_distinguish}. The \(1^{\text{st}}\)-order Maximum Mean Discrepancy (MMD) \cite{kernel_sequential_data, higher_order_kme_neurips} between stochastic processes is the distance between their respective \(1^{\text{st}}\)-order KMEs. The \(1^{\text{st}}\)-order MMD distinguishes between distributions of stochastic processes \cite{first_order_mmd_distinguish}. However, in certain cases, we need to distinguish between the conditional laws of stochastic processes. The authors in \cite{higher_order_kme_neurips} extend the \(1^{\text{st}}\)-order MMD to a higher-order MMD. Given two stochastic processes \(\mathbb{X}\) and \(\mathbb{Y}\), the \(2^{\text{nd}}\)-order MMD is defined as \(\mathcal{D}^{2} \left(\mathbb{X}, \mathbb{Y}\right) = \norm{\mu_{\pmb{X}}^{2} - \mu_{\pmb{Y}}^{2}}\). The \(2^{\text{nd}}\)-order MMD is a stronger discrepancy measure than the \(1^{\text{st}}\)-order MMD \cite[Theorem 2]{higher_order_kme_neurips}. \newline

As previously mentioned, we do not have access to the full distributions but instead, data samples are available. These samples are then used to calculate the distances using an unbiased and consistent \cite[Theorem 3]{higher_order_kme_neurips} empirical estimator. Suppose we have \(n\) sample paths \(\left\{\pmb{x}^{i} \in \right.\) \(\left. \mathcal{X} \left( \mathcal{K} \right) \right\}_{i=1}^{n}\) from \(\mathbb{X}\) and \(m\) sample paths \(\left\{\pmb{y}^{j} \in \mathcal{X} \left( \mathcal{K} \right) \right\}_{j=1}^{m}\) from \(\mathbb{Y}\). The samples \( \left\{\pmb{\tilde{x}}^{i} \right\}_{i=1}^{n} \sim \mu^{1}_{\pmb{X} | \mathbb{T}} \) and \( \left\{\pmb{\tilde{y}}^{j} \right\}_{j=1}^{m} \sim \mu^{1}_{\pmb{Y} | \mathbb{T}} \) are constructed (Appendix \ref{app:mmd}) and the empirical estimate is given by
\begin{equation} \label{eqn:mmd_2_estimate}
        \hat{\mathcal{D}}^{2} \left( \mathbb{X}, \mathbb{Y} \right)^{2} \coloneqq \frac{1}{n \left(n-1\right)} \sum_{\substack{i, j=1 \\
        i \neq j}}^{n} k_{\mathcal{S}} \left( \pmb{\tilde{x}}^{i}, \pmb{\tilde{x}}^{j} \right) - \frac{2}{nm} \sum_{i, j=1}^{n, m} k_{\mathcal{S}} \left( \pmb{\tilde{x}}^{i}, \pmb{\tilde{y}}^{j} \right) + \frac{1}{m \left(m-1\right)} \sum_{\substack{i, j=1 \\
        i \neq j}}^{m} k_{\mathcal{S}} \left( \pmb{\tilde{y}}^{i}, \pmb{\tilde{y}}^{j} \right).
\end{equation}

\section{A Scalable Solution to the Higher-Order DR Problem} \label{sec:approach}

Generally, the DR learning task is addressed using kernel methods. When the concept of DR on stochastic processes was initially introduced, the kernel used was a function of the \(1^{\text{st}}\)-order MMD. However, the DR function may be discontinuous with respect to the topology induced by the \(1^{\text{st}}\)-order MMD. By transitioning to higher-order distances, a finer topology than in the case of the \(1^{\text{st}}\)-order MMD is induced. Continuity of the DR function is achieved with respect to the finer topologies induced by the higher-order spaces. Recent work has shown that preserving continuity of the DR function translates to better performance in certain tasks \cite{higher_order_kme_neurips}. Applying higher-order distribution regression, the authors of \cite{higher_order_kme_neurips} use a kernel based on the \(2^{\text{nd}}\)-order MMD. This kernel was of the form \(k \left( \cdot, \cdot \right) = f \circ \mathcal{D}^{2} \left(\cdot, \cdot \right) \), for some appropriate function $f \colon \mathbb{R} \to \mathbb{R}$. For example, if \(f \left(x\right) = \exp\left(-x^{2}/2\sigma^{2}\right)\), then \(k\) is similar to the Radial Basis Function (RBF) kernel, however, in this case, instead of Euclidean norm between vectors, the squared \(2^{\text{nd}}\)-order MMD between stochastic processes is used. To reduce the computational overhead attributed to the KRR solution, we use a different regression technique and compute distances to a fixed number of reference sets (landmark points) \cite{conference_paper, reference_points_paper}. Although this significantly reduces the number of distance computations, using empirical estimators to compute the distances is still computationally challenging and infeasible when working with large batch sizes, large datasets, long time series and high-dimensional processes. \newline

To perform DR on stochastic processes, current state of the art approaches make use of KRR \cite{dist_reg_paper, higher_order_kme_neurips}. KRR is ideal for small datasets, but, applying it to large datasets is infeasible. Suppose we have \(N\) stochastic processes \(\mathbb{X}^{i}\) in our training dataset. Using KRR, we know that the optimal solution is of the form \(F \left(\cdot\right) = \sum_{i=1}^{N} \alpha_{i} k \left(\mathbb{X}^{i}, \cdot \right)\). To obtain a prediction for a new stochastic process \(\mathbb{Y}\), the kernel is evaluated \(N\) times. Suppose we have \(n\) \(d\)-dimensional sample paths from each stochastic process. Since the kernel \(k \left( \cdot, \cdot \right) \) is a function of the MMD between the stochastic processes, to compute the kernel, the solution to \(3n^{2} - 2n\) PDEs of the form in Eq.\ \eqref{eqn:goursat_pde} are needed to evaluate the \(1^{\text{st}}\)-order MMD. Transitioning to higher-order distances amplifies the computational issues since it necessitates significantly more computations. If the solution grid used to evaluate the PDE is of size $P^{2}$, computing the \(2^{\text{nd}}\)-order MMD has a time complexity of \(\mathcal{O} \left(dn^{2}P^{2} + n^{3}P^{2}\right)\) \cite{higher_order_kme_neurips}. Clearly, computing the \(2^{\text{nd}}\)-order MMD is time consuming since it is cubic in the number of samples and quadratic in the size of the solution grid. Hence, there is a tradeoff between accuracy of the PDE solutions and number of samples used. Also, computing \(N\) distances using empirical estimates required for evaluating the KRR solution compounds the computational problems, especially if certain compute resources are not available. In addition, once memory considerations are taken into account, a careful design choice is needed, taking into account the batch size $n$, the dimension of the time series, the length of the time series, the cardinality of the dataset, and the size of the solution grid. Consequently, current DR approaches are limited to tasks in which the batch size, dimension of the path, and the dataset are small. These limit the practicality of DR to various ML problems whilst also leading to training and testing instability. \newline

To reduce the computation time, the authors in \cite{conference_paper} train a neural network to map stochastic model parameters to distances. This facilitated the use of large datasets as well as neural networks to perform the regression. Moreover, their pre-trained \(2^{\text{nd}}\)-order MMD approximator removes all ambiguity in feature vectors since the output of the model is deterministic given the same stochastic model parameter pairs as input. This approach is ideal in situations when there is a strong level of confidence on the stochastic model class used to model the data. This is because, if the underlying distribution of the data is non-stationary, the pre-trained MMD approximator will need constant updating to reflect the new data regimes. If there is a regime change between training and testing, the stochastic models used during training might not be a good fit to the data in the new regime. In this case, a new MMD approximator needs to be trained to accommodate the new stochastic model class describing the new regime. To address these issues, we propose a novel \(2^{\text{nd}}\)-order MMD approximator which is based on sample paths as input. Our approximator input is independent of path length and batch size. Hence, larger batch sizes can be used. This stabilises the learning procedure by reducing errors in the empirical distribution approximating the true distribution. \newline

To avoid re-training an MMD approximator for each new dimension considered (i.e.\ an approximator for \(2\)-dimensional paths, a new approximator for \(3\)-dimensional paths, etc.\ ) we propose altering the methodology to be \(1\)-dimensional focused. We do this by changing the structure of the feature vectors from the ones used by the models in \cite{conference_paper} and \cite{dist_reg_paper}. In these works, for a $d$-dimensional stochastic process (\(d > 1\)), the MMD of the entire $d$-dimensional distribution is computed. Instead, we propose computing the MMD between marginal distributions, hence restricting the paths to \(1\)-dimensional paths. Consider the $d$-dimensional stochastic processes (\(d\geq1\)) \(\mathbb{X}^{1}, \cdots, \mathbb{X}^{N}\) and let \(\mathbb{X}_{j}^{i}\) denote the \(j^{\text{th}}\)-marginal of the \(i^{\text{th}}\) process. We propose regressing features of the form
\begin{equation*}
    \left( \mathcal{D}^{2} \left(\mathbb{X}_{1}^{i}, \mathbb{Z}_{1}^{1}\right)^{2}, \cdots, \mathcal{D}^{2} \left(\mathbb{X}_{1}^{i}, \mathbb{Z}_{1}^{B_{1}}\right)^{2}, \cdots, \mathcal{D}^{2} \left(\mathbb{X}_{d}^{i}, \mathbb{Z}_{d}^{1}\right)^{2}, \cdots, \mathcal{D}^{2} \left(\mathbb{X}_{d}^{i}, \mathbb{Z}_{d}^{B_{d}}\right)^{2} \right)
\end{equation*}
\noindent
where \(\mathbb{Z}_{j}^{p}\) denotes the \(j^{\text{th}}\) marginal of the \(p^{\text{th}}\) stochastic base process corresponding to the reference set specific to the \(j^{\text{th}}\) marginal. \(B_{j}\) denotes the cardinality of the reference set corresponding to the \(j^{\text{th}}\) marginal. The regression models, in our case neural networks, are able to extrapolate the co-dependence structure between marginals from the distances to the reference sets.

\section{Model-agnostic 2nd-order MMD approximator} \label{sec:mmd_approx_2}

Since the empirical estimator of the MMD computes distances using paths sampled from both stochastic processes, we propose a solution based on sample paths rather than stochastic model parameters. Under certain assumptions, the expected signature \(\mathbb{E} \left[ \mathcal{S} \left( \mathbb{X} \right) \right]\) characterises the distribution of \(\mathbb{X}\) \cite{unique_sig_dist}. We encode the distributions using expected signatures. Since the signature is an infinite-dimensional tensor, when approximating the expected signature, we need to use the truncated signature. If we set a high level of truncation, we would be prioritising information over computation time. On the other hand, by setting a low level of truncation we might miss out on important information. \newline

Consider two stochastic processes \(\mathbb{X}, \mathbb{Y}\). We sample \(n\) paths \(\left\{\pmb{x}^{i} \in \right.\) \(\left. \mathcal{X} \left( \mathcal{K} \right) \right\}_{i=1}^{n} \sim \mathbb{X}\) and \(m\) paths \(\left\{\pmb{y}^{j} \in \mathcal{X} \left( \mathcal{K} \right) \right\}_{j=1}^{m} \sim \mathbb{Y}\). The inputs to the neural network approximating the squared \(2^{\text{nd}}\)-order MMD are of the form \(\left(\mathbb{E} \left[ \mathcal{S} \left(\mathbb{X} \right) \right], \mathbb{E} \left[ \mathcal{S} \left(\mathbb{Y} \right) \right] \right)\). To compute the \(2^{\text{nd}}\)-order MMD, we lift the paths from the ambient space to a feature map using a static kernel \(\phi\) \cite{sig_kernel_pde_paper}. In our case, we set \(\phi\) to be the RBF kernel. When constructing the inputs to the neural network, we compute the signature of kernelised paths \(\Psi \left(\pmb{x}^{i} \right) = \exp\left(- \left(\pmb{x}^{i} \right)^{2} \right)\). Let \(L\) denote the level of truncation. In this case, the input to the neural network is
\begin{equation*}
\left(\mathbb{E} \left[ \mathcal{S} \left(\Psi \left(\mathbb{X} \right) \right) \right], \mathbb{E} \left[ \mathcal{S} \left( \Psi \left(\mathbb{Y} \right) \right) \right] \right) \approx \left( n^{-1} \sum_{i=1}^{n} \mathcal{S}^{L} \left( \Psi \left[ \pmb{x}^{i} \right] \right), m^{-1} \sum_{j=1}^{m}  \mathcal{S}^{L} \left( \Psi \left[\pmb{y}^{j} \right] \right) \right).
\end{equation*}
The \(2^{\text{nd}}\)-order MMD captures conditional distribution properties of the stochastic processes. Since we approximate the \(2^{\text{nd}}\)-order MMD using the expected signature, the neural network could be acting as a smoothing function allowing it to extrapolate the information needed to approximate the correct level of MMD. Although there might be situations where the filtration is not being captured (either fully or partially), by computing distances to multiple reference sets, this is mitigated when performing DR.

\subsection{Training}\label{sec:mmd_approx_exp}

To train our \(2^{\text{nd}}\)-order MMD approximator, we generate sample paths using three stochastic model classes encompassing a broad range of dynamical systems. These are:
\begin{itemize}[nosep,left=0pt]
    \item The Geometric Brownian Motion (GBM) \cite{black_scholes_paper}. This stochastic model is described by the Stochastic Differential Equation (SDE)
    \begin{equation*}
        dx_{t}^{\text{gbm}} = \mu x_{t}^{\text{gbm}} dt + \sigma x_{t}^{\text{gbm}} dW_{t}
    \end{equation*}
    \noindent
    where \(\left(W_{t}\right)\) is a standard Brownian Motion, \(\mu, \sigma > 0\).
    \item A mean reverting model with stochastic volatility \cite{heston_model_paper}. The dynamics are described by the SDEs
    \begin{equation*}
dx_{t}^{\text{mr}} = \mu x_{t}^{\text{mr}}dt + \sqrt{\sigma_{t}}x_{t}^{\text{mr}}dW^{x}_{t}, ~~d\sigma_{t} = \kappa \left( \theta - \sigma_{t} \right) dt + \xi \sqrt{\sigma_{t}} dW_{t}^{\sigma}
\end{equation*}
\noindent
where \(\left(W^{x}_{t}\right)_{t}, \left(W^{\sigma}_{t}\right)_{t}\) are correlated Brownian Motions with correlation \(\rho\) and \(\mu, \kappa, \theta, \xi > 0\).
\item The rough Bergomi model (rBergomi) \cite{pricing_rough_vol, fclt, turbo}. In this model, the randomness is driven by a fractional Brownian Motion. The dynamics are described by the equation 
\begin{equation*}
    x_{t}^{\text{rb}} = x_{0}^{\text{rb}} \mathcal{E} \left( \int_{0}^{t} \sqrt{\sigma_{u}} dW_{u}^{x}  \right),~~ \sigma_{u} = \xi_{0}  \mathcal{E} \left( \nu \sqrt{2H} \int_{0}^{u} \left(u-s\right)^{H - 1/2} dW_{s}^{\sigma}\right)
\end{equation*}
\noindent
where \(\mathcal{E} \left( \cdot \right) \) is the stochastic Wick exponential \cite{wick_exp}, \( \xi_{0} > 0, \nu > 0, H \in \left(0, 0.5\right)\), and \(\left(W^{x}_{t}\right)_{t}, \left(W^{\sigma}_{t}\right)_{t}\) are correlated Brownian Motions with correlation \(\rho\). \newline

\end{itemize}

The training dataset consisted of sample paths from all three stochastic model classes (simulation details can be found in Appendix \ref{app:numerical_simulation}). Since the \(2^{\text{nd}}\)-order MMD is a metric, the dataset was augmented to include \(10{,}000\) distances with a value of \(0\). These correspond to distances between two identical stochastic models. This was done to ensure the structure of a metric was preserved by the neural network. In total, the dataset was of size \(39{,}449\). A neural network was trained to minimise Mean Squared Error (MSE). Model training and validation performance compared across various truncation levels are presented in Table \ref{tab:mmd_validation_mse}. We also verified that this neural network approximates a metric. Additional model and training details can be found in Appendix \ref{app:mmd_approx}. \newline

An ablation study was conducted to investigate the effect of including \(10{,}000\) samples with zero distance. The training and validation MSE of the model trained without zero distances is comparable to the model trained with zero distances included. We then checked whether the model outputs a zero distance when the same stochastic model is used as input (\(\text{Model} \left( \mathbb{X}, \mathbb{X} \right) \approx 0\)). The model trained without zero distances satisfied the condition \(\text{Model} \left( \mathbb{X}, \mathbb{X} \right) < 0.1\) in \(31\%\) of the 10,000 test samples, whereas the model trained with zero distances satisfied the condition in \(88\%\) of test cases. This indicates that the model needs to be trained with zero distances to preserve the metric structure of the MMD. The tests described in this section took approximately \(1\) hour to run on a GTX 1660 Ti GPU.

\begin{table}
  \caption{Average MMD approximator MSE. Standard deviations across \(5\) independent model runs.}
  \label{tab:mmd_validation_mse}
  \centering
  \begin{tabular}{ccccc}
    \toprule
    Trunc. (\(L\))  & \(\left|\mathcal{S}^{L} \left(\cdot \right) \right|\) & Hidden Layer Dim. & Train. MSE (\(\times 10^{-3}\))    &  Valid. MSE (\(\times 10^{-3}\)) \\
    \midrule
    \(2\)     & \(12\) & \(25\) & \(13.2\) (\(\pm 0.14\))      & \(13.7\) (\(\pm 0.46\))  \\
    \(3\) &  \(28\) & \(60\) & \(4.5\) (\(\pm 0.061\))  & \(5.4\) (\(\pm 0.28\))     \\
    \(4\)     & \(60\) & \(90\) & \(2.7\) (\( \pm 0.28 \)) & \(4.2\) (\( \pm 0.53\))     \\
    \bottomrule
  \end{tabular}
\end{table}

\section{Experiments}\label{sec:exp}

We tested our approach on \(3\) different tasks; one from the physical sciences, one from estimation theory, and the other from quantitative finance. We also tested the generalisation capabilities of our model in out-of-sample distributional tests. All distances were computed using level \(3\) truncation.

\paragraph{Baselines} Our methodology was benchmarked against \(3\) baselines. The baselines used are similar to the ones adopted by \cite{dist_reg_paper}. We used DeepSets \cite{deep_sets_paper} as one of the benchmarks. In this model, the data is treated as a permutable set. The DeepSets model consists of two neural networks. The first neural network processes the items in the set individually. It then aggregates the outputs, and feeds this aggregated output as input to the neural network responsible for performing the regression. The DeepSets model is sensitive to model architecture choices and activation functions \cite{deep_sets_2, dist_reg_paper}. The other two baseline models are kernel based approaches. We adopted our DR methodology and used pointwise kernels instead of the signature kernel. Neural networks were then used to perform the regression. Two pointwise kernels were considered; the RBF kernel and the Matern32 kernel. Additional details on the kernel-based baselines are available in Appendix \ref{app:exp}. \newline

Besides the DeepSets neural networks, all other neural networks (SPEEDRS and baselines) consisted of \(3\) hidden layers each with an activation function and an output layer with linear activation function. Each neural network had a fixed hidden layer dimension. This was reported in Table \ref{tab:pricing_validation_mse}. The neural networks were trained using MSE with L2-regularisation. Gradient updates were performed using AdamW \cite{adamW_paper}. To train each model, the dataset was randomly split into a training and validation set according to the ratio \(80{:}20\) \cite{dataloader}. Except for the DeepSets model, input data was standardised.

\subsection{Mixture Models}

Mixture models \cite{mixture_models_review, gausian_mixture_models} describe probability distributions composed of a weighted sum of known densities. They provide a flexible framework for modelling unknown distributional shapes and have been used for classification, inference, and modelling complex surfaces \cite{mixture_3, mixture_2}. They feature across multiple domains including health sciences \cite{mixture_application_1}, life sciences \cite{mixture_application_2}, and finance \cite{mixture_3, mixture_2, mixture_1, mixture_4, mixture_5}. 

\paragraph{Training} Stochastic processes were modelled as a mixture between a mean reverting process and a rBergomi process. The mixture process is given by \(x_{t} = \alpha x_{t}^{\text{mr}} + \left(1-\alpha \right)x_{t}^{\text{rb}}\) with \(\alpha\) being the mixture parameter. Reference sets (reference models) \(\left\{ \mathbb{X}_{i} \right\}_{1 \leq i \leq N}\) were split among the rBergomi, mean reverting, and GBM models according to the ratio \(2{:}2{:}1\). Reference sets were randomly selected. To address the estimation uncertainty in the empirical distribution, batch sizes of \(2{,}000\) sample paths were used. This was made possible by using our MMD approximator. Due to computational constraints, the kernel baseline features were constructed using batch sizes of \(400\) sample paths.

\subsubsection{Derivative Pricing} \label{sec:pricing_exp}

We tested our methodology on pricing down-and-in barrier options where the payoff is triggered if the running minimum of the asset price time series goes below a pre-set barrier level (\(B\)) by the expiry time (\(T\)). The payoff of this derivative is path-dependent and discontinuous at the barrier level. These make it challenging to price, and modelling of the trajectory is crucial for accurate pricing. Mathematically, this derivative is described by the payoff \(\max \left(x_{T} - K, 0\right) \mathbf{1}_{m_{T} \leq B} \)
where \(m_{T} \coloneqq \min_{0 \leq u \leq T} x_{u}\) denotes the running minimum up to maturity. \(\mathbf{1}_{\left(\cdot\right)}\) denotes the indicator function and \(K\) is the strike level. We obtained the (target) prices using Monte Carlo pricing \cite{mc_mgbm_book, mc_methods_finance_book}. The training data consisted of prices corresponding to models with mixture parameter values \(\alpha = 0, 0.25, 0.5, 0.75, 1.0\). The MSEs averaged across \(5\) independent runs are reported in Table \ref{tab:pricing_validation_mse}. SPEEDRS outperforms all baselines considered. The DeepSets model achieved comparable training MSE to SPEEDRS, however it performed significantly worse on the validation set. To the best of our knowledge, SPEEDRS is the first model for pricing derivatives under more than one stochastic model class. \newline

\begin{figure}
     \centering
     \begin{subfigure}{0.45\textwidth}
         \centering
         \includegraphics[width=0.98\linewidth, height=0.25\textheight]{ 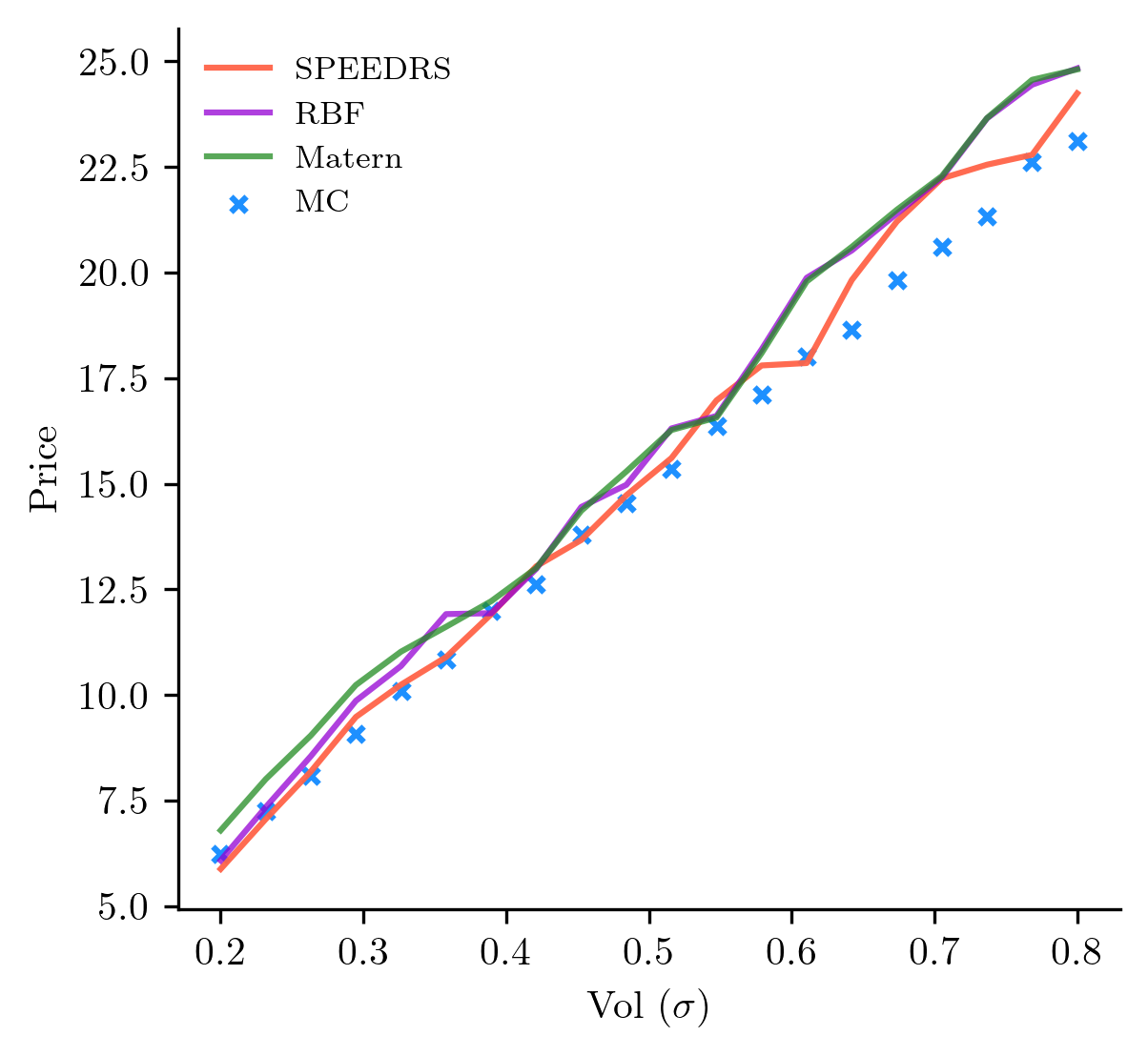}
         \caption{GBM.}
         \label{fig:gbm_prices}
     \end{subfigure}
    \begin{subfigure}{0.45\textwidth}
         \centering
         \includegraphics[width=0.98\linewidth, height=0.25\textheight]{ 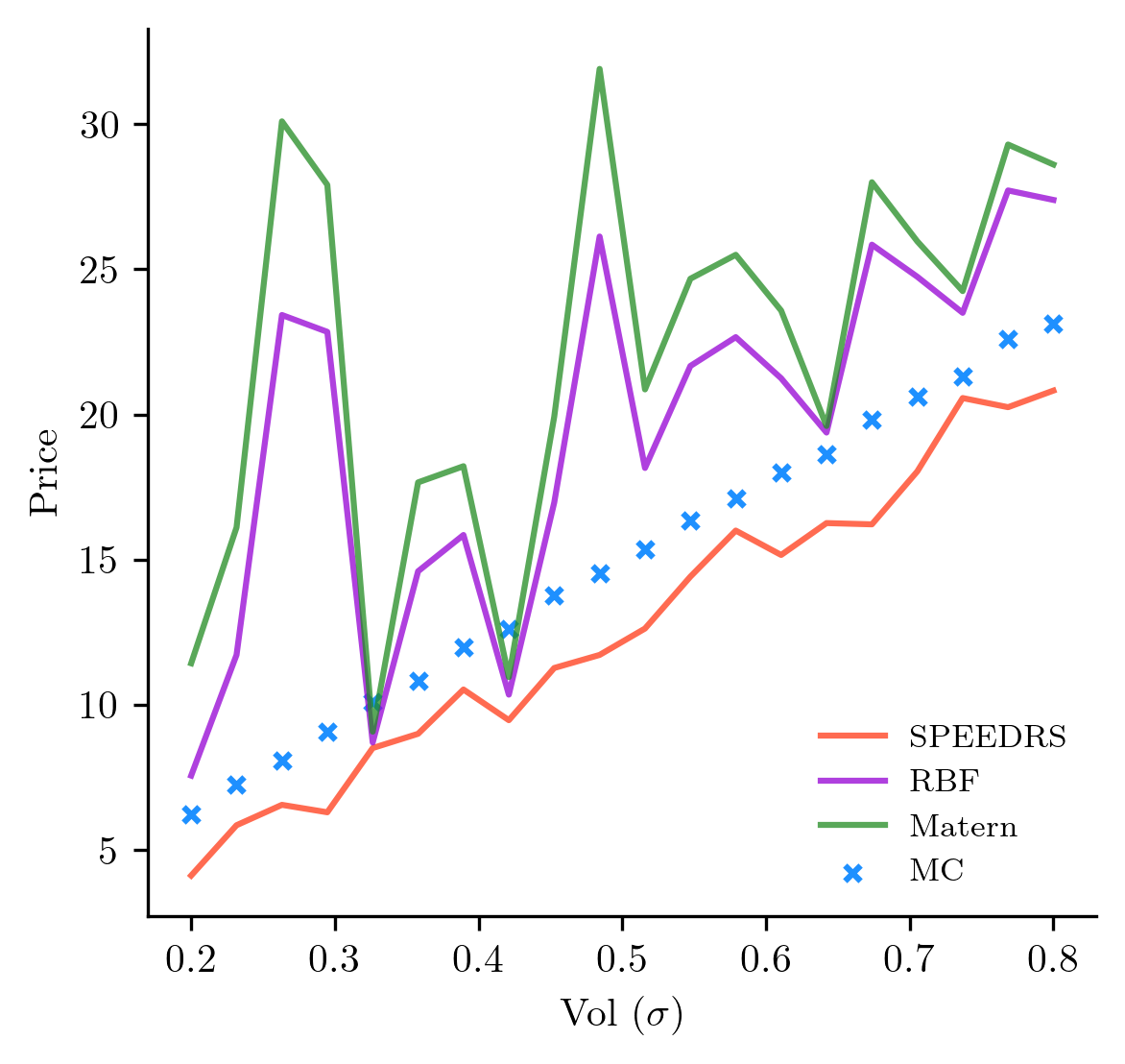}
         \caption{GBM.}
         \label{fig:gbm_prices_irr}
     \end{subfigure}
     \begin{subfigure}{0.45\textwidth}
         \centering
         \includegraphics[width=0.98\linewidth, height=0.25\textheight]{ 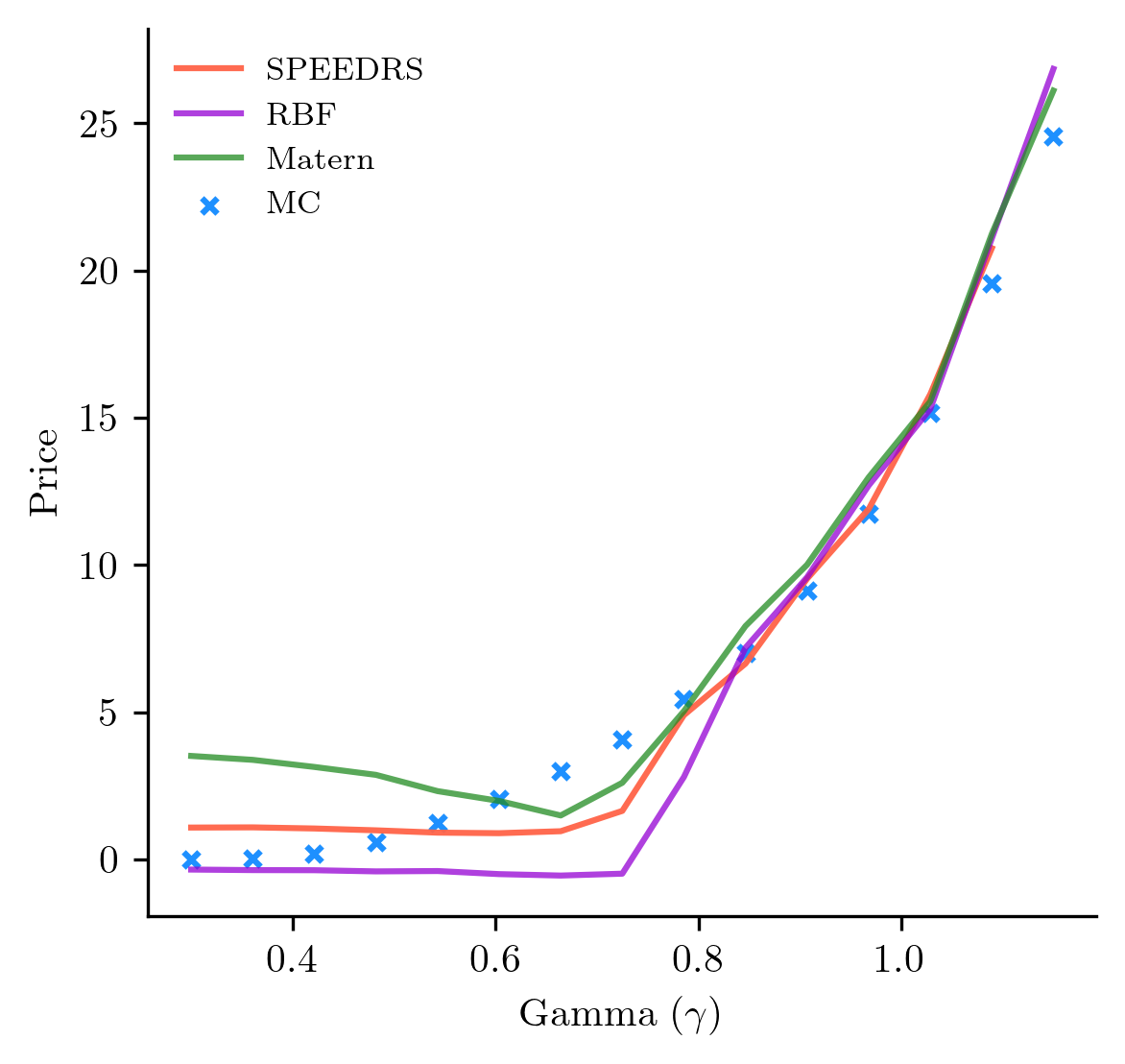}
         \caption{CEV.}
         \label{fig:cev_prices}
     \end{subfigure}
     \begin{subfigure}{0.45\textwidth}
         \centering
         \includegraphics[width=0.98\linewidth, height=0.25\textheight]{ 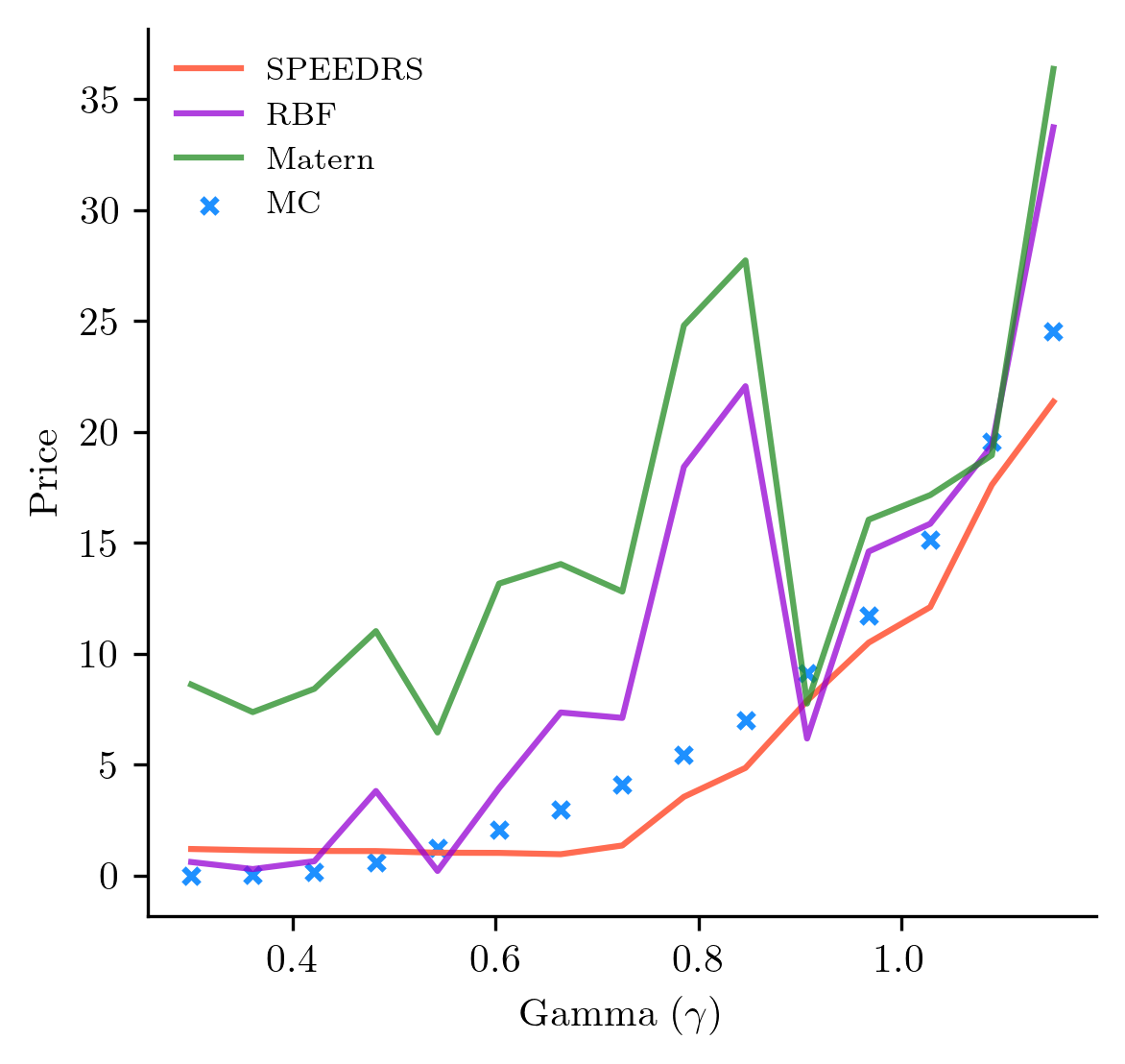}
         \caption{CEV.}
         \label{fig:cev_prices_irr}
     \end{subfigure}
    \caption{Pricing under unseen regimes. (a), (c) regular sampling. (b), (d) irregular sampling.}
    \label{fig:oos_prices}
\end{figure}

The out-of-sample tests were targeted at stress-testing the distributional robustness of our methodology. All tests were carried out with \(20\) reference sets and results compared with the RBF and Matern baselines. Two aspects of model drift were assessed; performance on unseen data within a given distribution and performance under unseen regimes. We further tested the robustness of SPEEDRS by using irregularly sampled data. To test the performance on unseen data within the same mixture setting, we tested the pricing model on mixture parameter values within the entire range \(\alpha \in \left[0, 1\right]\) instead of restricting the mixture parameter values used during training. SPEEDRS obtained a MSE of \(0.05\), where each predicted price was the average of \(100\) independent runs through SPEEDRS. To test the model under unseen regimes, we generated prices using the GBM model and the Constance Elasticity of Variance (CEV) model \cite{cev_paper}, two classes of stochastic models which did not feature in the training dataset. The CEV model is a generalisation of the GBM with dynamics described by \(dx_{t}^{\text{cev}} = \mu x_{t}^{\text{cev}} dt + \sigma \left(x_{t}^{\text{cev}}\right)^{\gamma} dW_{t}\). When pricing under these two model classes, we also ran tests in which we irregularly subsample our time series. All results are provided in Figure \ref{fig:oos_prices}. Clearly, SPEEDRS outperforms the RBF and Matern baseline models, achieving good generalisation performance in unseen regimes whilst also performing relatively well under irregularly sampled time series data from these unseen regimes. This could be as a result of using higher-order signatures as well as the use of \(4.5\) times the amount of samples to approximate the true distribution in the case of our signature approach. All experiments took approximately \(2\) hours to run on a GTX 1660 Ti GPU. 

\subsubsection{Mixture Parameter Estimation}

\begin{figure}
     \centering
     \begin{subfigure}{0.45\textwidth}
         \centering
         \includegraphics[width=0.98\linewidth, height=0.25\textheight]{ 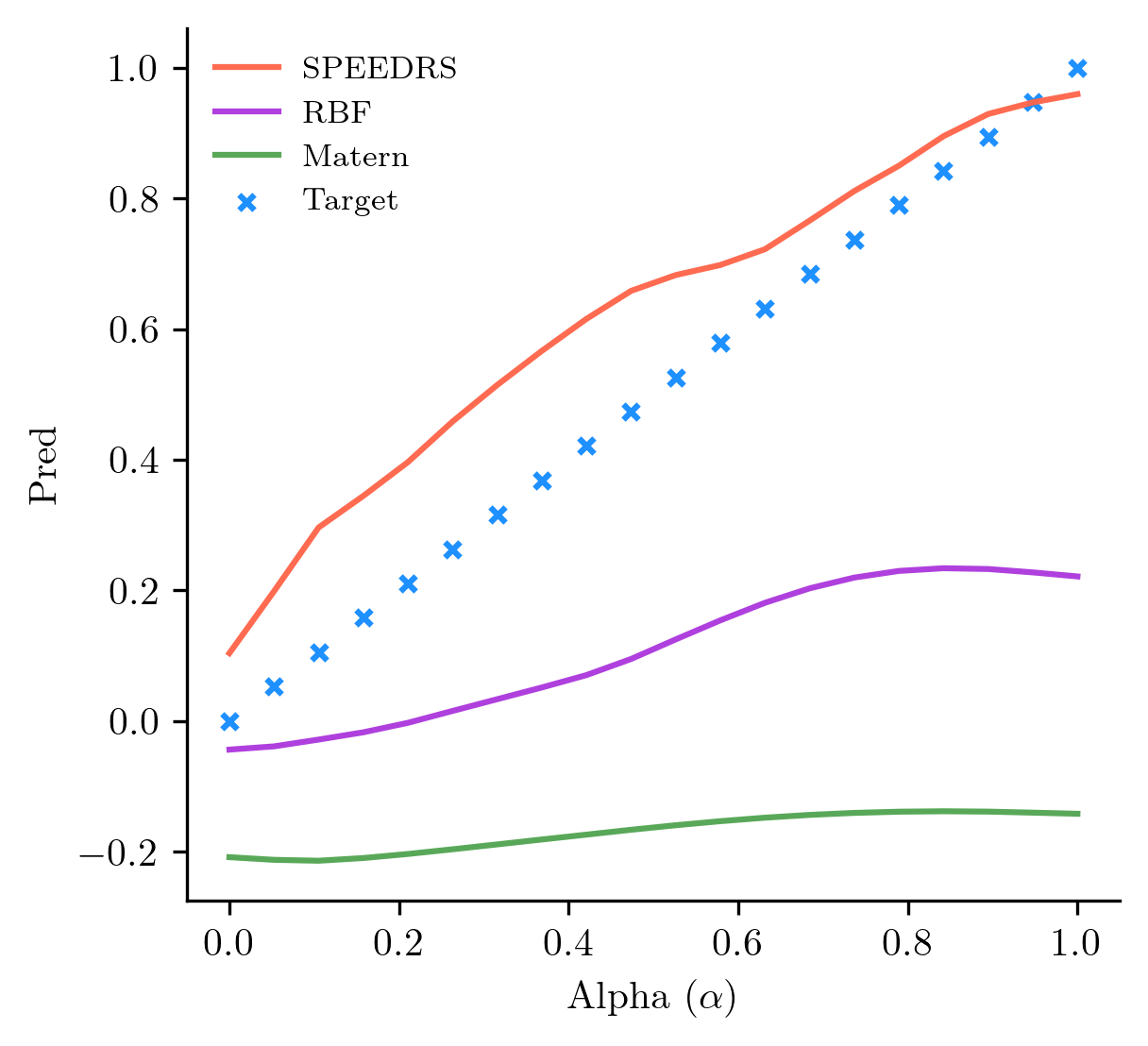}
         \caption{rBergomi.}
         \label{fig:bregomi_param}
     \end{subfigure}
    \begin{subfigure}{0.45\textwidth}
         \centering
         \includegraphics[width=0.98\linewidth, height=0.25\textheight]{ 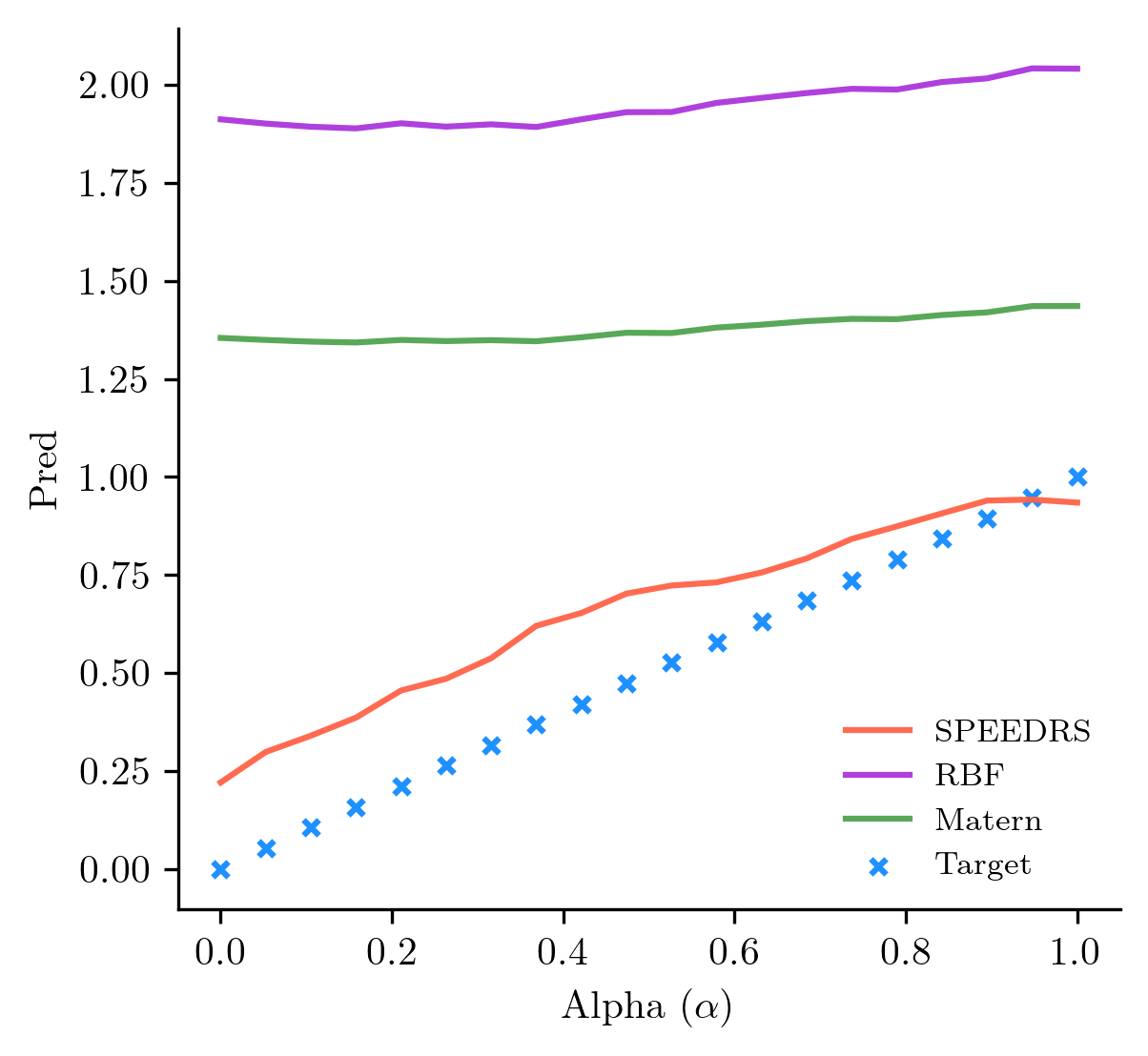}
         \caption{rBergomi.}
         \label{fig:bregomi_param}
     \end{subfigure}
     \begin{subfigure}{0.45\textwidth}
         \centering
         \includegraphics[width=0.98\linewidth, height=0.25\textheight]{ 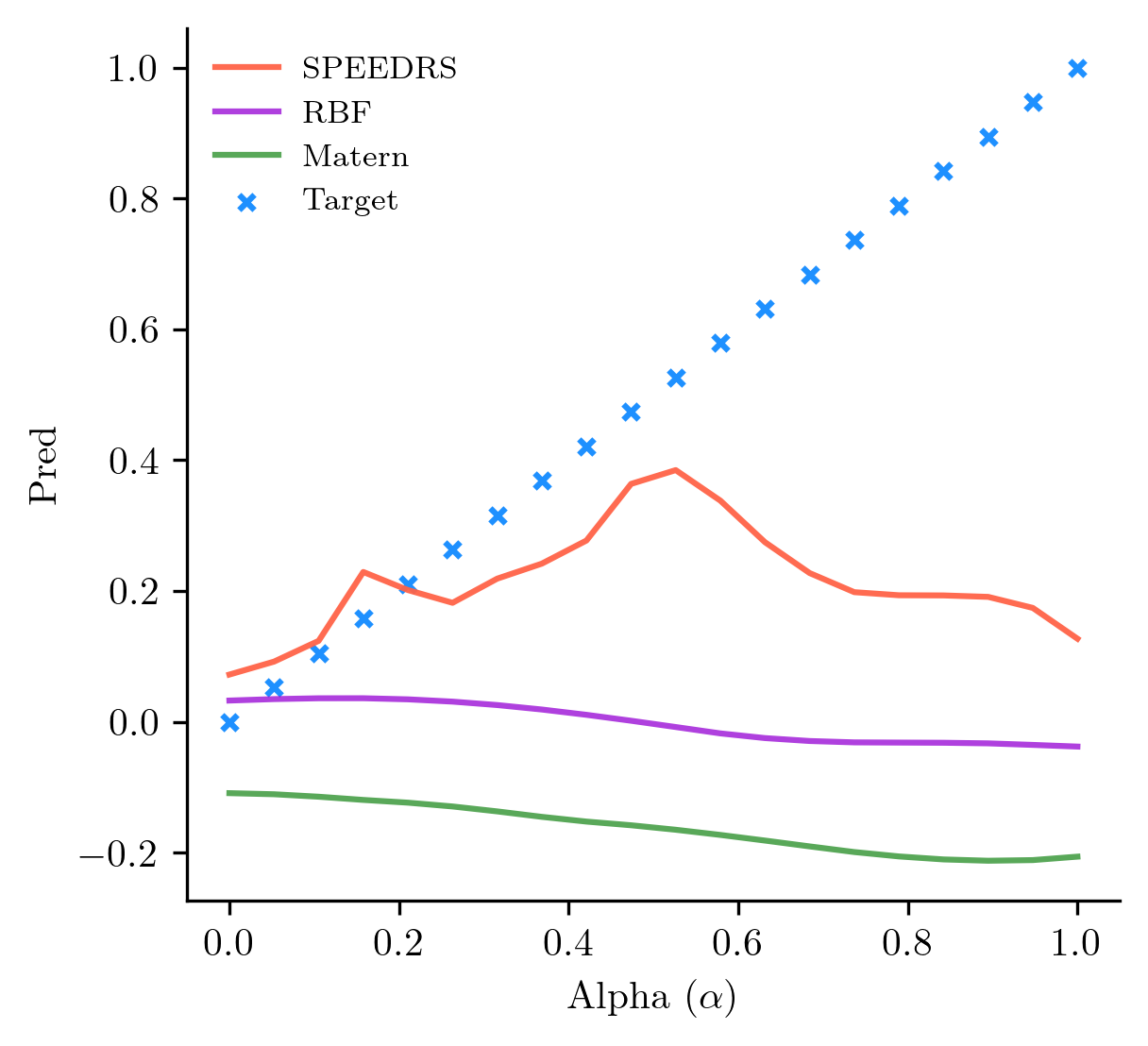}
         \caption{Mean reverting.}
         \label{fig:heston_param}
     \end{subfigure}
    \begin{subfigure}{0.45\textwidth}
         \centering
         \includegraphics[width=0.98\linewidth, height=0.25\textheight]{ 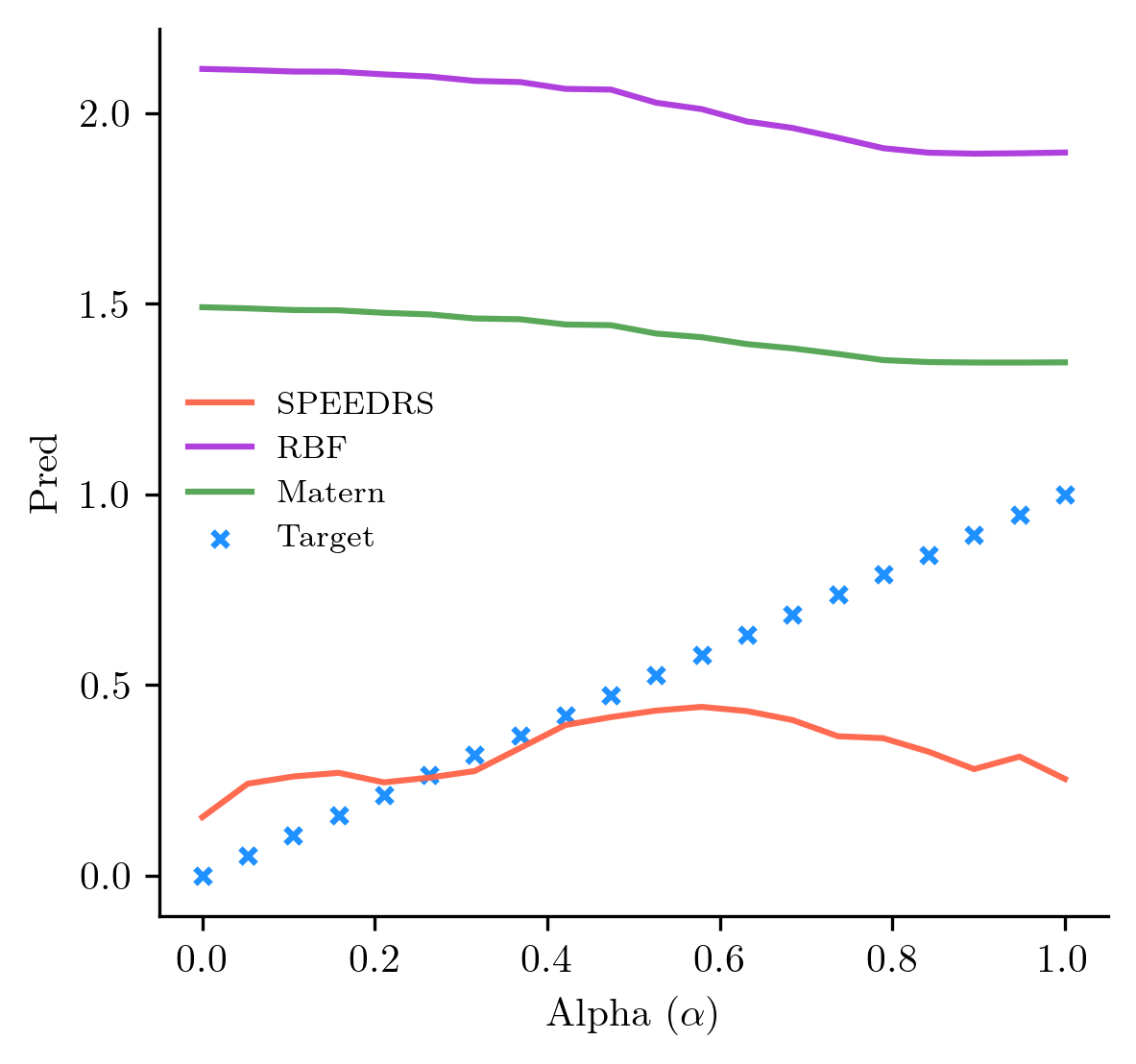}
         \caption{Mean reverting.}
         \label{fig:heston_param}
     \end{subfigure}
    \caption{Mixture parameter estimation under unseen regimes. (a), (c) regular sampling. (b), (d) irregular sampling.}
    \label{fig:oos_param}
\end{figure}

A challenging task when working with mixture models is estimating the mixture parameter given observed data. This involves approportioning the data across different distributions, which in our case consists of a non-Markov process (rBergomi) and a Markov one. Since one stochastic model is path-dependent whilst the other is not, this increases the complexity of the problem. To train the models, the mixture parameter was sampled uniformly within the range \(\left[0, 1\right]\). Results of SPEEDRS and the baseline models are provided in Table \ref{tab:pricing_validation_mse}. SPEEDRS outperformed all baselines in terms of training and validation MSE. To test the out-of-sample performance of SPEEDRS, we once again tested the model under regime changes. We tested the performance of the model when estimating the mixture parameter between the mixture pairs (rBergomi, CEV) and (CEV, Mean reverting). These pairs did not feature in the training dataset. As was done for the derivative pricing experiment, the performance under irregularly sample data was also tested. The results are depicted in Figure \ref{fig:oos_param}. Once again, SPEEDRS outperforms the baselines. However, in the second mixture test, the performance of SPEEDRS, although significantly better than the baselines, drops past a certain mixture parameter value.

\subsection{Temperature of an Ideal Gas}

We consider the task of inferring the temperature of an ideal gas from particle movements within a cube. A challenge posed by this problem is that the time series are multi-dimensional. This increases the complexity of the model. The thermodynamic properties of the gas inside a cube of volume \(V\) are described through the temperature (\(T\)), the pressure (\(P\)), and the energy (\(U\)). These three properties are related through the equations \cite{dist_reg_paper} \(PV = Nk_{\text{B}}T\) and \(U=c_{V}Nk_{\text{B}}T\), where \(k_{\text{B}}\) is the Boltzman-constant, and \(c_{V}\) is the heat capacity \cite{boltzman_constant}. The large-scale behaviour of the gas is related to the trajectories of the particles\footnote{Trajectories are computed using the code available at \url{https://github.com/labay11/ideal-gas-simulation}.} and therefore depends on the temperature. Another factor influencing the dynamics is the radii of the particles, where for a fixed temperature, the larger the radii the higher the chance of a collision between particles. For a fixed volume \(V\) and batch size \(n\), we simulate particles of radii \(r = 0.35 \left(V/n\right)^{1/3}\). We set the batch size to \(800\) and trajectory length to \(20\). A dataset of \(5{,}000\) samples was used. Model results are reported in Table \ref{tab:pricing_validation_mse}. Due to  high-dimensionality of the problem, the DeepSets baseline was not tested.

\begin{table}
  \caption{Average model MSE. Standard deviations and model dimensions are also included. \(B\) denotes the number of reference sets.}
  \label{tab:pricing_validation_mse}
  \centering
  \resizebox{\textwidth}{!}{\begin{tabular}{cccccc}
    \toprule
    Application & Model & \(B\)  & NN Layer Dim. & Train. MSE (\(\times 10^{-1}\))    &  Valid. MSE (\(\times 10^{-1}\)) \\
    \midrule
    \midrule
    \multirow{6}{2em}{Pricing} & SPEEDRS & \(5\) & \(15\) & \(10.4\) (\(\pm 0.39\)) & \(10.6\) (\(\pm 0.39\))    \\
    & SPEEDRS & \(10\)  & \(30\) & \(5.5\) (\(\pm 0.11\)) & \(6.0\) (\(\pm 0.25\))     \\
    & SPEEDRS & \(20\) & \(50\)   & \(\mathbf{4.6}\) \textbf{(}\(\mathbf{\pm 0.08}\)\textbf{)}  & \(\mathbf{5.1}\) \textbf{(}\(\mathbf{\pm 0.21}\)\textbf{)}  \\
    \cline{3-6} 
    & \rule{0ex}{2ex} RBF & \(20\) & \(50\) & \(11.1\) (\(\pm 0.33\)) & \(11.7\) \(\pm 0.55\) \\
    & Matern & \(20\) & \(50\) & \(11.5\) (\(\pm 0.13\)) & \(11.6\) (\(0.55\)) \\
    & DeepSets & N/A & N/A & \(5.9\) (\(\pm 0.46\)) & \(1267.9\) (\(\pm 1.2\)) \\
    \midrule
    \multirow{6}{2em}{Estim.} & SPEEDRS & \(5\) & \(15\) & \(0.094\) (\(\pm 0.0032\)) & \(0.095\) (\(\pm 0.0048\))    \\
    & SPEEDRS & \(10\)  & \(30\)   & \(0.071\) (\(\pm 0.0022\)) & \(0.072\) (\(\pm 0.0034\))     \\
    & SPEEDRS & \(20\) & \(50\)   & \(\mathbf{0.052}\) \textbf{(}\(\mathbf{\pm 0.0023}\)\textbf{)}  & \(\mathbf{0.056}\) \textbf{(}\(\mathbf{\pm 0.0024}\)\textbf{)}  \\
     \cline{3-6} 
    & \rule{0ex}{2ex} RBF & \(20\) & \(50\) & \(0.064\) (\(\pm 0.0004\)) & \(0.066\) (\(\pm 0.0017\)) \\
    & Matern & \(20\) & \(50\) & \(0.064\) (\(\pm 0.0012\)) & \(0.065\) (\(\pm 0.0018\)) \\
    & DeepSets & N/A & N/A & \(0.11\) (\(\pm 0.006\)) & \(0.11\) (\(\pm 0.0051 \)) \\
    \midrule
    \multirow{6}{2em}{Temp} & SPEEDRS & \(9\) & \(50\) & \(0.88\) (\(\pm 0.051\)) & \(1.19\) (\(\pm 0.077\))    \\
    & SPEEDRS & \(15\)  & \(75\)   & \(0.54\) (\(\pm 0.057\)) & \(0.70\) (\(\pm 0.058\))     \\
    & SPEEDRS & \(30\) & \(100\)   & \(\mathbf{0.45}\) \textbf{(}\(\mathbf{\pm 0.081}\)\textbf{)} & \(\mathbf{0.61}\) \textbf{(}\(\mathbf{\pm 0.047}\)\textbf{)}  \\
    \cline{3-6} 
    & \rule{0ex}{2ex} RBF & \(30\) & \(100\) & \(3.6\) (\(\pm 0.12\)) & \(3.4\) (\(\pm 0.17\)) \\
    & Matern & \(20\) & \(50\) & \(3.7\) (\(\pm 0.12\)) & \(3.7\) (\(\pm 0.33\)) 
  \end{tabular}}
\end{table}

\section{Limitations}

As mentioned above, the MMD approximator uses the expected signature whilst the \(2^{\text{nd}}\)-order MMD is based on higher-order signatures, capturing additional properties than what is captured by the expected signature. By computing distances to multiple reference sets, this drawback is mitigated when performing DR. However, this needs to be investigated further if the approximator is to be used for other purposes, such as two-sample hypothesis testing \cite{first_order_mmd_distinguish}. As noted by the authors of \cite{mmd_break} and \cite{higher_order_kme_neurips}, if a more generic set of continuous sample paths than \(\mathcal{X} \left( \mathbb{R}^{d} \right) \) is considered, the regularity of the paths might not propagate to the higher-order paths. If this is the case, the KMEs will not be well defined. Our work does not account for these paths.

\section{Conclusion and future work}

In this paper we propose a methodology which directly addresses the computational barriers present within the higher-order DR framework. We achieve this by using a higher-order DR solution based on distances to reference sets. These distances are calculated using our novel \(2^{\text{nd}}\)-order MMD approximator. We demonstrate that our methodology is applicable to a broad range of ML tasks and show that SPEEDRS performs well under distributional changes. Areas for possible future work include researching the optimal collection of reference sets for a given problem, and a distance approximator ideal for two-sample hypothesis testing. 

\section*{Acknowledgments}

This work was supported by the UK Engineering and Physical Sciences Research Council (EPSRC) under Grant [EP/T517963/1].

\bibliographystyle{plainurl}
\bibliography{mmd_pricing_references}

\appendix

\section{Empirical Estimator of the \(\mathbf{2}^{\text{nd}}\)-Order MMD} \label{app:mmd}

 Suppose we have \(n\) sample paths \(\left\{\pmb{x}^{i} \in \right.\) \(\left. \mathcal{X} \left( \mathcal{K} \right) \right\}_{i=1}^{n}\) from \(\mathbb{X}\) and \(m\) sample paths \(\left\{\pmb{y}^{j} \in \mathcal{X} \left( \mathcal{K} \right) \right\}_{j=1}^{m}\) from \(\mathbb{Y}\). The empirical estimate for the \(2^{\text{nd}}\)-order MMD is given by
 \begin{equation*} 
        \hat{\mathcal{D}}^{2} \left( \mathbb{X}, \mathbb{Y} \right)^{2} \coloneqq \frac{1}{n \left(n-1\right)} \sum_{\substack{i, j=1 \\
        i \neq j}}^{n} k_{\mathcal{S}} \left( \pmb{\tilde{x}}^{i}, \pmb{\tilde{x}}^{j} \right) - \frac{2}{nm} \sum_{i, j=1}^{n, m} k_{\mathcal{S}} \left( \pmb{\tilde{x}}^{i}, \pmb{\tilde{y}}^{j} \right) + \frac{1}{m \left(m-1\right)} \sum_{\substack{i, j=1 \\
        i \neq j}}^{m} k_{\mathcal{S}} \left( \pmb{\tilde{y}}^{i}, \pmb{\tilde{y}}^{j} \right).
\end{equation*}
\noindent
To compute this distance, we need to evaluate the signature kernel between conditional KMEs. This computation is based on covariance and cross-covariance operators \cite{cross_covariance_1, cross_covariance_2, cross_covariance_3}. Let \(\mathbf{I}_{m}\) denote the \(m \times m\) identity matrix. Suppose \(\tilde{\pmb{x}} \sim \mu_{\mathbf{X} | \mathbb{T}}^{1}\) and \(\tilde{\pmb{y}} \sim \mu_{\mathbf{Y} | \mathbb{T}}^{1}\). For \(0 \leq s, t \leq T\), 
\begin{equation*}
    k_{\mathcal{S}} \left(\tilde{\pmb{x}}_{s}, \tilde{\pmb{y}}_{t} \right) = \langle\tilde{\pmb{x}}_{s}, \tilde{\pmb{y}}_{t} \rangle = \left(\mathbf{k}_{s}^{\pmb{x}}\right)^{\intercal} \left(\mathbf{K}_{s, s}^{\pmb{x}, \pmb{x}} + n \lambda \mathbf{I}_{n} \right)^{-1}\mathbf{K}_{T, T}^{\pmb{x}, \pmb{y}} \left(\mathbf{K}_{t, t}^{\pmb{y}, \pmb{y}} + m \lambda \mathbf{I}_{m} \right)^{-1}\mathbf{k}_{t}^{\pmb{y}}
\end{equation*}
where \(\mathbf{k}_{s}^{\pmb{x}} \in \mathbb{R}^{n}\) and \(\mathbf{k}_{t}^{\pmb{y}} \in \mathbb{R}^{m}\) are the vectors
\begin{equation*}
    \left[ \mathbf{k}_{s}^{\pmb{x}} \right]_{i} \coloneqq k_{\mathcal{S}} \left(\pmb{x}^{i}_{\left[0, s\right]}, \pmb{x}_{\left[0, s\right]} \right)~~\text{and}~~\left[ \mathbf{k}_{t}^{\pmb{y}} \right]_{i} \coloneqq k_{\mathcal{S}} \left(\pmb{y}^{i}_{\left[0, t\right]}, \pmb{y}_{\left[0, t\right]} \right)
\end{equation*}
\noindent
and \(\mathbf{K}_{s, s}^{\pmb{x}, \pmb{x}} \in \mathbb{R}^{n \times n}, \mathbf{K}_{t, t}^{\pmb{x}, \pmb{y}} \in \mathbb{R}^{m \times m}, \mathbf{K}_{T, T}^{\pmb{x}, \pmb{y}} \in \mathbb{R}^{n \times m}\) are the matrices
\begin{equation*}
    \left[\mathbf{K}_{s, s}^{\pmb{x}, \pmb{x}}\right]_{i, j} \coloneqq k_{\mathcal{S}} \left(\pmb{x}_{\left[0, s\right]}^{i}, \pmb{x}_{\left[0, s\right]}^{j}\right),~ \left[\mathbf{K}_{t, t}^{\pmb{x}, \pmb{y}} \right]_{i, j} \coloneqq k_{\mathcal{S}} \left(\pmb{y}_{\left[0, t\right]}^{i}, \pmb{y}_{\left[0, t\right]}^{j} \right),~\left[\mathbf{K}_{T, T}^{\pmb{x}, \pmb{y}}\right]_{i, j} \coloneqq k_{\mathcal{S}} \left( \pmb{x}_{\left[0, T\right]}^{i}, \pmb{y}_{\left[0, T\right]}^{j} \right). 
\end{equation*}
Full details of the algorithm are provided in \cite{higher_order_kme_neurips}. 

\section{Numerical Simulation} \label{app:numerical_simulation}

To compute distances between stochastic processes, we simulated \(M\) sample paths over the time horizon \(\left[0, T\right]\). This was done by discretising \(\left[0, T\right]\) into \(n+1\) equally spaced time points. Let \(dt = T/n\). We discretised \(\left[0, T\right]\) by constructing the grid \(\mathbb{T}^{n} \coloneqq \left\{t_{i} \right\}_{i=0}^{n}\) where 
\begin{equation*}
    t_{i} = i \frac{T}{n} = idt.
\end{equation*}

The GBM was simulated using the Euler-Maruyama method. For \(i = 1, \cdots, n\), the update rule is given by
\begin{equation*}
    S_{t_{i}} = S_{t_{i-1}} + \mu S_{t_{i-1}} dt + \sigma S_{t_{i-1}} \left(W_{t_{i}} - W_{t_{i-1}} \right) = S_{t_{i-1}} + \mu S_{t_{i-1}} dt + \sigma \sqrt{dt} S_{t_{i-1}} Z,~~S_{0}>0
\end{equation*}
\noindent
where \(Z \sim \mathcal{N} \left(0, 1\right)\) is a standard normal random variable since \(\left(W_{t}\right)\) is a Brownian motion. In the case of the CEV model, the update rule is
\begin{equation*}
    S_{t_{i}} = S_{t_{i-1}} + \mu S_{t_{i-1}} dt + \sigma \sqrt{dt} S_{t_{i-1}}^{\gamma} Z,~~S_{0}>0,~~i=1, \cdots, n.
\end{equation*}

To simulate the mean reverting model, we simulate both the volatility process \(v_{t}\) and the price process \(S_{t}\). In our simulations, we updated the price process before the volatility. Also, to avoid negative volatilities when computing the square root, we computed the square root of \(\max \left(0, v_{t} \right)\). The simulation scheme is given by
\begin{align*}
    &S_{t_{i}} = S_{t_{i-1}} + \mu S_{t_{i-1}} dt + \sqrt{\max \left(v_{t_{i-1}}, 0 \right)}  \sqrt{dt} S_{t_{i-1}} Z^{S}, &&& \\\nonumber
    &v_{t_{i}} = v_{t_{i-1}} + \kappa \left(\theta - v_{t_{i-1}} \right) dt + \xi \sqrt{dt} \sqrt{\max \left(v_{t_{i-1}}, 0 \right)} \left(\rho Z^{S} + \sqrt{1 - \rho^{2}} Z^{v} \right),~~i=1, \cdots, n
\end{align*}
\noindent
where \(Z^{S}, Z^{v}\) are independent standard normal random variables.

To simulate the rBergomi process, we followed the work in \cite{fclt}. We simulated the log-price process. The log-dynamics are given by
\begin{align*}
    dX_{t} &= -\frac{1}{2} V_{t} dt + \sqrt{V_{t}} dW_{t}^{X},~~X_{0} = \log \left(S_{0}\right) &&& \\\nonumber
    v_{t} &= \Phi \left(\phi\right) \left(t \right),~~v_{0} > 0 &&& \\\nonumber
    \Phi \left(\phi \right) \left(t\right) &= \xi_{0} \exp \left(2 \nu \sqrt{2H} \phi \left(t\right) \right) \exp \left(-4 \nu^{2} H \int_{0}^{t} \left(t-s\right)^{2H - 1} ds \right) &&& \\\nonumber
    \phi \left(t\right) &= \int_{0}^{t} \left(t-s\right)^{H - 1/2} dW_{s}^{v}.
\end{align*}
\noindent
To simulate the log-dynamics, we adapted Algorithm 3.3 of \cite{fclt}. In our simulations, we used antithetic variates to reduce variance \cite{fclt}. This was done as follows:

Let \(\left\{ \zeta_{i, j} \right\}_{i, j=1}^{n, M/2}\) and \(\left\{ \xi_{i, j} \right\}_{i, j=1}^{n, M/2}\) be two collections of independent standard normal random variables. Set \(Z^{v} \coloneqq \left\{ \zeta_{i, j} \right\}_{i, j=1}^{n, M/2} \cup \left\{ -\zeta_{i, j} \right\}_{i, j=1}^{n, M/2}\) and \(Z^{S} \coloneqq \left\{ \rho \xi_{i, j} + \sqrt{1- \rho^{2}} \zeta_{i, j} \right\}_{i, j=1}^{n, M/2} \cup \left\{ \rho \xi_{i, j} - \sqrt{1- \rho^{2}} \zeta_{i, j} \right\}_{i, j=1}^{n, M/2}\). Define \(g \left(u\right) \coloneqq u^{H - 1/2}\) and \(\mathbf{g} \coloneqq \left\{g \left(t_{i}^{\ast}\right) \right\}_{i=1}^{n}\) where \(t_{i}^{\ast}\) are the optimal moment matching evaluation points 

\begin{equation*}
    t_{i}^{\ast} \coloneqq \left(\frac{n}{2H} \left[ \left(t - t_{i-1} \right)^{2H} - \left(t - t_{i}\right)^{2H} \right] \right)^{\frac{1}{2H-1}}
\end{equation*}
\noindent
provided in \cite[Section 3.3.1]{fclt}. Let \(\left(\phi\right)^{j} \left(\mathbb{T}^{N}\right)\) denote the \(j^{\text{th}}\) discretised path of \(\phi \left(t\right)\) over \(\mathbb{T}^{N}\). Then
\begin{equation*}
    \left(\phi\right)^{j} \left(\mathbb{T}^{N}\right) = \sqrt{\frac{T}{n}} \left( \mathbf{g} \ast Z^{v}_{j} \right),~~~j = 1,\cdots,M
\end{equation*}
\noindent
where \(\ast\) is the discrete convolution operator. For each \(i=1, \cdots, n\) and \(j = 1, \cdots, M\), the log-dynamics are simulated using the formula
\begin{align*}
    X^{j} \left(t_{i}\right) &= X^{j} \left(t_{0}\right) - \frac{1}{2} dt \sum_{k=1}^{i} \Phi \left(\phi \right)^{j} \left(t_{k-1}\right) + \sqrt{dt} \sum_{k=1}^{i} \sqrt{ \Phi \left(\phi \right)^{j} \left(t_{k-1}\right)} Z^{S}_{k, j} && \\\nonumber
    &= X^{j} \left(t_{i-1}\right) - \frac{1}{2} dt \Phi \left(\phi \right)^{j} \left(t_{i-1}\right) + \sqrt{dt} \sqrt{ \Phi \left(\phi \right)^{j} \left(t_{i-1}\right)} Z^{S}_{i, j}.
\end{align*}
This simulation scheme was used to simulate all rBergomi models besides the models forming part of the reference sets in the parameter estimation and derivative pricing experiment. In this case, we simulated the conditional log-price process. Let \(\mathcal{F}_{t}^{W^{X}}, \mathcal{F}_{t}^{W^{v}}\) be the natural filtrations generated by the Brownian motions \(\left(W_{t}^{X}\right), \left(W_{t}^{v}\right)\) respectively. As mentioned in \cite{turbo, cond_rbergomi}, the conditional process  
\begin{equation*}
    X_{t} | \left( \mathcal{F}_{t}^{W^{v}} \vee \mathcal{F}_{0}^{W^{X}} \right)
\end{equation*} 
\noindent
is normally distributed. The simulation of the conditional log-price process is very similar to the one for the unconditional dynamics. To simulate the conditional log dynamics, we adapted Algorithm 3.5 in \cite{fclt}.

The code used to simulate the mean reverting model was adapted from \cite{github_heston_ref} and the simulation procedure and code for the rBergomi model was adapted from \cite{fclt}. To generate the training data, stochastic model parameters were sampled uniformly within the ranges
\begin{itemize}
    \item \(\xi_{0} \in \left[0.01, 0.2\right) \)
    \item \(\nu \in \left[0.5, 4.0 \right)\)
    \item \(H \in \left[0.025, 0.5\right)\)
    \item \(v_{0}, \theta, \kappa, \xi, \sigma \in \left[0.2, 0.8\right)\)
    \item \(\rho \left[-1, 1\right)\)
    \item \(\mu \in \left[0.01, 0.2\right)\).
\end{itemize}

\section{MMD Approximator Training Details} \label{app:mmd_approx}

\paragraph{Dataset} Each stochastic process was represented using \(400\) sample paths of length \(15\) and all paths were normalised to start at \(1\). The \(2^{\text{nd}}\)-order MMDs between processes were computed using the code provided in \cite{higher_order_kme_neurips} and with static kernel set to the RBF kernel with \(\sigma=0.5\). All distances were calculated using A\(100\) GPUs. Running the code to construct the dataset took approximately \(3.5\) hours.  

\begin{figure}
         \centering
         \includegraphics[width=0.6\linewidth, height=0.15\textheight]{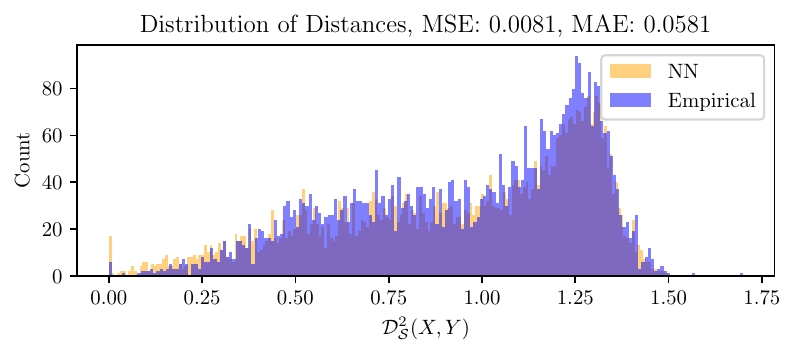}
         \caption{Distribution of target distances (blue) and predictions (yellow).}
         \label{fig:mmd_approx_2_distances}

\end{figure}

\paragraph{Training} We trained a neural network consisting of three hidden layers with ReLU activation function and an output layer with linear activation function. Inputs were standardised and a learning rate decay \cite{lr_decay_paper} was used on an initial learning rate of \(5 \times 10^{-4}\). All models were trained with mini-batches \cite{deep_learning_book} using the AdamW optimiser \cite{adamW_paper} over \(200\) epochs on a GTX \(1660\) Ti GPU. The model was trained to minimise the MSE with L\(2\)-regularisation having hyperparameter \(1 \times 10^{-4}\). To train the model, the dataset was randomly split into a training (\(80\%\)) and validation (\(20\%\)) set \cite{dataloader}. \(5\) separate models were trained. Training all models took approximately \(45\) minutes. In Figure \ref{fig:mmd_approx_2_distances}, we plot a histogram of the distances in our test set and the distances predicted by our model. The MSE of these predictions was \(8.1 \times 10^{-3}\) and the mean absolute error was \(5.81 \times 10^{-2}\).

\section{Experiments} \label{app:exp}

\subsection{Baselines}

The baseline kernels used are the RBF kernel and the Matern32 kernel. These are defined as follows
\begin{align*}
    &k_{\text{RBF}}^{\sigma} \left(\pmb{x}_{i}, \pmb{x}_{j}\right) \coloneqq \exp \left(-\frac{\norm{\pmb{x}_{i} - \pmb{x}_{j}}^{2}}{2 \sigma^{2}} \right), && \\\nonumber
    &k_{\text{Matern32}}^{\sigma} \coloneqq \left(1 + \sqrt{3}\sigma^{2} \norm{\pmb{x}_{i} - \pmb{x}_{j}}\right) \exp \left( -\sqrt{3}\sigma^{2} \norm{\pmb{x}_{i} - \pmb{x}_{j}}^{2} \right).
\end{align*}
 In our experiments, we set \(\sigma=1\). RBF and Matern32 baselines, we use the following procedure. Let \(\left\{\pmb{x}_{i}\right\}_{i \leq N}\) denote a collection of \(N\) \(d\)-dimensional paths of length $l$. To construct features, the time series are first flattened into an \(Nd \times l\) matrix. Then, the \(l\)-dimensional distribution of the vectors is embedded in a RKHS with kernel \(k_{\text{baseline}}\). A reference set of distributions is used to calculate distances between distributions. Given \(B\) reference sets \(\pmb{z}^{i} = \left\{\pmb{z}^{i}_{j}\right\}\), features of the form
\begin{equation*}
    \left\{N^{-1} \sum_{i, j=1}^{N} k_{\text{baseline}} \left(\pmb{x}_{i}, \pmb{x}_{j}\right) - 2N^{-2} \sum_{i, j=1}^{N} k_{\text{baseline}} \left(\pmb{x}_{i}, \pmb{z}^{k}_{j}\right) +  N^{-1} \sum_{i, j=1}^{N} k_{\text{baseline}} \left(\pmb{z}^{k}_{i}, \pmb{z}^{k}_{j}\right) \right\}_{k=1}^{B}.
\end{equation*}
are regressed against scalar targets using neural networks.

\subsection{Derivative Pricing Experiment Training Details}

\paragraph{Dataset} To construct the dataset, we fixed \(x_{0} = 90, K = 80, B = 85\) and we varied the rBergomi stochastic model parameters, mean reverting process parameters, and mixture parameter \(\alpha\). In our training dataset, we set \(\alpha \in \left\{0.25 \cdot i~|~i = 0, \cdots, 4 \right\}\). The dataset was of size \(10{,}000\) and required \(2\) hours of computation time to construct.

\begin{figure}
         \centering
         \includegraphics[width=0.6\linewidth, height=0.2\textheight]{ 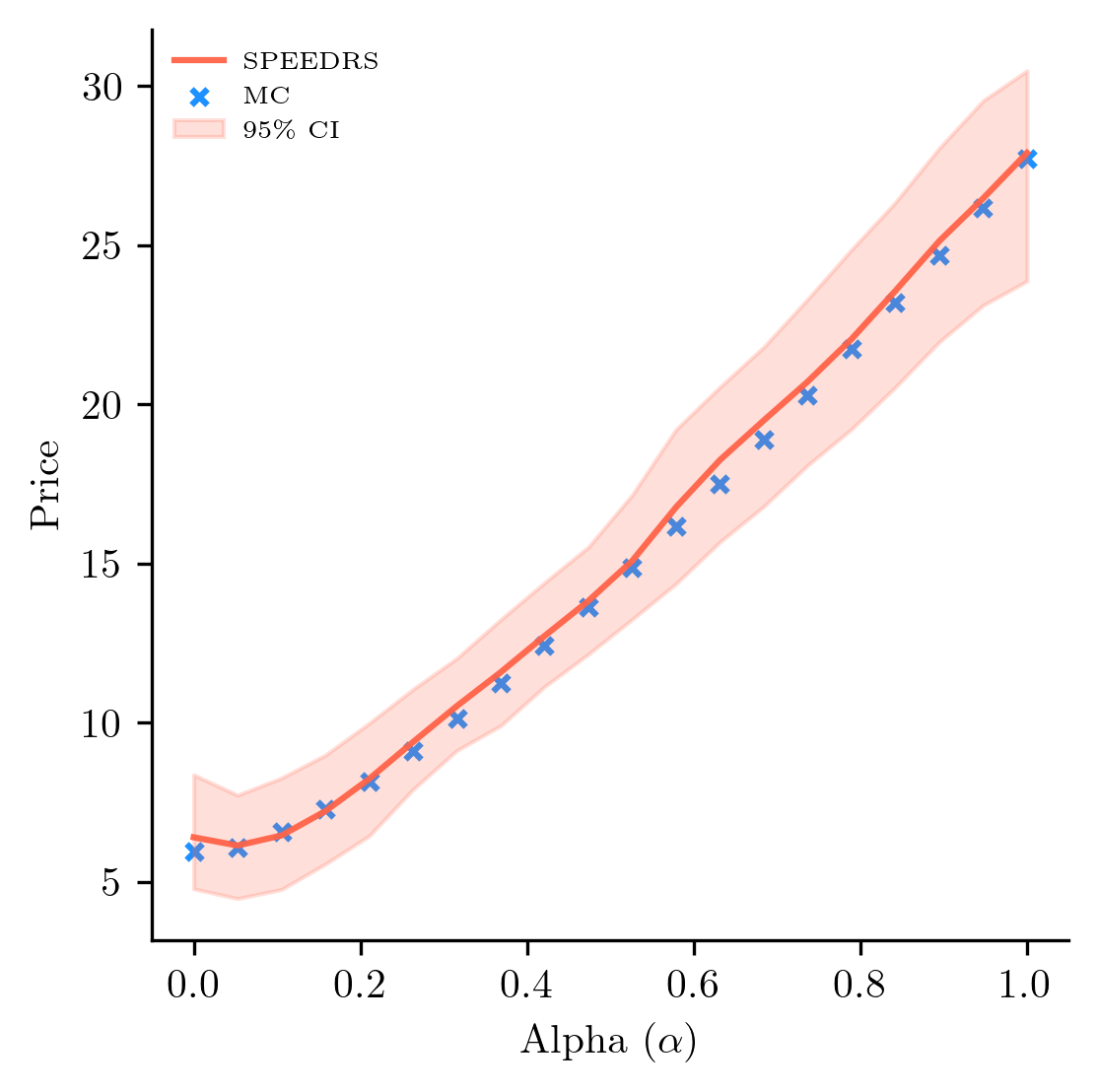}
         \caption{Barrier option prices under a mixture of Heston and rBergomi with \(\mu = 0.15, v_{0}^{\text{Heston}} = 0.65, \theta = 0.23, \kappa = 0.4, \xi^{\text{Heston}} = 0.25, \rho^{\text{Heston}} = -0.94, v_{0}^{\text{rBergomi}} = 0.75, H = 0.25, \nu = 1.2, \xi_{0} = 0.1, \rho^{\text{rBergomi}} = -0.85\).}
         \label{fig:mixt_prices}

\end{figure}

\paragraph{SPEEDRS}  The neural networks were trained over \(200\) epochs. The L\(2\)-regularisation hyperparameter used was \(1 \times 10^{-4}\). ReLU activation was used. The initial learning rate was set to \(5 \times 10^{-4}\) and a learning rate decay \cite{lr_decay_paper} was used. Models were trained using the AdamW \cite{adamW_paper} optimiser on a GTX 1660 Ti GPU. Training lasted around \(25\) minutes. An out-of-sample test showing the robustness of SPEEDRS to distributional shifts is depicted in Figure \ref{fig:mixt_prices}

\paragraph{Baselines} Generating the RBF and Matern32 features took approximately \(1\) hour combined. The neural networks were identical to the ones used for SPEEDRS. The DeepSets model was trained on an A100 GPU using \(500\) epochs. Training lasted around 1.5 hours and a learning rate of \(1 \times 10^{-2}\) was used. 

\subsection{Mixture Parameter Estimation Training Details}

\paragraph{Dataset} The dataset was constructed by simulating mean reverting and rBergomi processes. The starting value was kept fixed at \(1\) and the terminal time was also fixed at \(1\). The paths were of length \(15\) and \(2{,}000\) paths were sampled from each distribution. Stochastic model parameters were sampled uniformly from the ranges presented in Appendix \ref{app:numerical_simulation}. Mixture parameter values were sampled uniformly in the range \(\left(0, 1\right)\). The mixture parameter values \(0, 1\) were included for every stochastic model parameter considered. The dataset consisted of \(14{,}000\) samples. Generating the dataset lasted less than \(0.5\) hours. 

\paragraph{SPEEDRS} Identical architecture, run-times, learning rate, and hyper-parameters as in the case of the derivative pricing task.

\paragraph{Baselines} Identical architecture, run-times, learning rate, and hyper-parameters as in the case of the derivative pricing task.

\subsection{Inferring Temperature of Ideal Gas Training Details}

\paragraph{Dataset} The dataset contained \(5{,}000\) stochastic processes each represented by \(800\) sample paths of length \(20\). The paths were of dimension \(3\) since the particles move around within a cube. Temperatures were sampled between \(1\) and \(10\). 

\paragraph{SPEEDRS} The neural networks were trained over \(200\) epochs. The L\(2\)-regularisation hyperparameter used was \(1 \times 10^{-4}\). Tanhshrink activation functions were used. The initial learning rate was set to \(1 \times 10^{-2}\) and a learning rate decay \cite{lr_decay_paper} was used. Models were trained using the AdamW \cite{adamW_paper} optimiser on a GTX 1660 Ti GPU. Training lasted around \(25\) minutes.

\paragraph{Baselines} Identical architecture, and hyper-parameters to the ones used for SPEEDRS. The only difference being the learning rate, which in this case was set to \(5 \times 10^{-4}\). Since the time series are of a higher dimension than the other tasks, constructing the baseline kernel features took \(2\) hours. Training of the baseline models was performed on a GTX 1660 Ti GPU and lasted under \(20\) minutes.

\end{document}